\documentclass{article}

\usepackage{amsmath,amsfonts,bm}

\def\eqref#1{equation~\ref{#1}}

\def\1{\bm{1}}

\def\vx{{\bm{x}}}

\def\mI{{\bm{I}}}

\DeclareMathAlphabet{\mathsfit}{\encodingdefault}{\sfdefault}{m}{sl}
\SetMathAlphabet{\mathsfit}{bold}{\encodingdefault}{\sfdefault}{bx}{n}

\usepackage{hyperref}
\usepackage{url}
\usepackage{amsmath,amssymb,amsthm,mathtools, natbib}
\usepackage{enumitem}
\usepackage{algorithm}
\usepackage{algorithmic}
\usepackage{float}
\usepackage{booktabs}
\usepackage{adjustbox}
\usepackage{makecell}
\usepackage[section]{placeins}
\usepackage{graphicx,authblk}
\usepackage{xcolor}
\usepackage{caption}
\newcommand{\asm}{ASM}
\newcommand{\asmid}{ASM-I}      
\newcommand{\asmsph}{ASM-S}     
\newcommand{\oraclek}{Oracle-$K_g$}
\newcommand{\fedgem}{FedGEM}

\usepackage{pifont}
\newcommand{\cmark}{\ding{51}}  
\newcommand{\xmark}{\ding{55}} 

\theoremstyle{plain}

\theoremstyle{remark}

\title{Federated Clustering with Unknown Local\\and Global Cluster Cardinalities}

\author{
  Mitushi Goyal, Tarun S., Riddhanya Senapathi, Arun Raman \\
  \small BITS Pilani K. K. Birla Goa Campus, Goa, India. 
}
\date{}
\begin{document}

\maketitle

\begin{abstract}
Federated clustering methods that do not require the global number of
clusters $K$ still assume that each client knows its local number $K_g$. This
assumption is hard to justify when clients know no more about their data than
the server does, as in fault diagnosis across independently operated
industrial sites. We propose a two-phase framework in which neither count is
known: each client first estimates $K_g$ from its own data, and an aggregator
that requires local counts, such as FedGEM, then uses these estimates in place
of the true values. For the first phase we introduce Adaptive Split--Merge
(ASM), which grows a spherical Gaussian mixture by BIC-driven splitting and
then merges excess components. ASM uses no labels, selects its hyperparameters
on held-out client data only, and makes no assumption about how clusters are
shared across clients. We derive a closed-form split criterion whose critical
cluster size falls with anisotropy and rises with dimension, and show
empirically that over-fragmentation grows with the number of points per
cluster, which federation divides among clients. Across eight datasets, ASM
with FedGEM attains a mean ARI of 0.333, against 0.256 for the next best
label-free estimator and 0.361 when the true local counts are supplied. It
also gives the most reliable global estimates of $K$ and is robust when client
size is decoupled from local cardinality.
\end{abstract}

\section{Introduction}
\label{sec:intro}

Federated clustering recovers cluster structure from unlabeled data held by
clients that cannot pool it for reasons of privacy, regulation or cost. Each
client can find structure in its own data, but the server must decide which of
these locally discovered structures correspond to the same global cluster,
without ever seeing the data behind them.

A concrete instance arises for original equipment manufacturers (OEMs) of
capital-intensive industrial systems such as power generators, which are
contractually bound to guarantee the reliability of equipment operated by their
clients \citep{ibrahim2026fedgem}. Meeting these guarantees requires the OEM to
detect and diagnose faults from sensor data collected across client sites, and
the problem is unsupervised and federated by necessity. The OEM does not know
every fault class in advance. Labels supplied by clients are unreliable because
labeling standards and maintenance practices differ from site to site. Raw
sensor data cannot be shared. FedGEM \citep{ibrahim2026fedgem} addresses this
setting without requiring the global number of clusters $K$, but it still
assumes that each client knows its own local number $K_g$. However, it is reasonable to assume that the plant operators know no more about the equipment's fault classes than the OEM that designed it, so the reasons $K$ is unknown apply equally to every $K_g$.

The same independence constrains how any method can be tuned. Each site
operates its equipment under its own maintenance regime and has no access to
other sites' data, so hyperparameters cannot be tuned jointly, either across
sites or between a site and the OEM. Sites can, however, exchange information
with the OEM at the level of parameters and summaries, repeatedly if needed.
This naturally separates the problem into two phases: each site first decides
on its own how finely to describe its data, and the OEM then reconciles these
descriptions across sites through repeated exchange of parameters.

The granularity with which a site describes its data matters in two ways. Clients can get their
own class-counts wrong, and what the server can do about it depends on the direction of the error. If a client reports two components where one was present, the server can merge them from the reported parameters alone. If it reports one where two were present, no correction is possible: many different pairs of components share the same mean and covariance, and resolving the ambiguity
would require a change in the information sharing structure. That argues for
reporting finely, but the same quantity governs disclosure: a site reporting a
single component reveals little beyond the gross statistics of its operation,
while one reporting a component per measurement has shared the raw data it was
contractually unable to send. The number of reported components therefore sits
on a continuum, with recoverability increasing along it and confidentiality
decreasing, and the first phase must place itself deliberately between the two.
Finally, how many points a site holds may or may not reflect how many clusters
it holds, depending on how its data were collected (for example, periodic logging, event-triggered logging, curated logging such as during diagnostic tests etc.), so an estimator of $K_g$
must not rely on client size as a proxy.

We propose a two-phase framework in which each client first estimates $K_g$
from its own data and a federated aggregator that requires local counts, such
as FedGEM, then takes these estimates in place of the true values. For the first phase we introduce the Adaptive Split--Merge (ASM) estimator, which grows a spherical Gaussian mixture (and thus, ASM-S) by recursive binary splitting: each candidate split is initialized along the principal direction, as in G-means
\citep{hamerly2003gmeans}, fitted by EM, and accepted when it improves the Bayesian Information Criterion, as in X-means \citep{pelleg2000xmeans}. ASM adds a merge stage that undoes excess splits, and selects all hyperparameters on held-out client data, without labels.

\paragraph{Contributions.}
\begin{itemize}
  \item \textbf{A two-phase framework for federated clustering in which neither
  the local nor the global number of clusters is known.} The framework is
  modular in both phases: any centralized method that estimates the number of
  clusters can serve as the local estimator, and any federated aggregator that
  requires local counts, such as FedGEM, can serve as the second phase, without
  modification to either.
  \item \textbf{ASM, a label-free local cardinality estimator} that combines
  BIC-driven splitting with a merge stage and selects its hyperparameters on
  each client's own held-out data (Section~\ref{sec:method}).
  \item \textbf{An analysis of when BIC splitting fragments a single cluster.}
  A closed-form criterion gives a critical cluster size that falls with
  anisotropy and rises with dimension. Synthetic experiments confirm it once
  small-sample inflation of anisotropy is accounted for and show that, beyond
  the critical size, splitting continues until the minimum-mass constraint
  binds. On pooled benchmarks, sample size governs the over-fragmentation of
  the split stage, and a bound derived from class anisotropy predicts where
  merging begins on seven of eight datasets (Section~\ref{sec:method}).
  \item \textbf{A federated evaluation that decouples client size from local
  cardinality.} Across eight datasets, ASM-S with FedGEM attains a mean ARI of
  0.333, against 0.256 for the next best label-free method and 0.361 when the
  true local counts are supplied. It gives the most reliable global estimates,
  with $|\hat K - K^*| \le 21$ against up to 188 and 263 for DP-GMM and G-means,
  and is unaffected when client size is made independent of $K_g$
  (Section~\ref{sec:experiments2}).
\end{itemize}

\section{Related work}
\label{sec:related}

\begin{table}[!t]
\centering
\caption{Clustering methods by the cardinality information each requires.
\dag~AFCL does not take $K_g$, but its initial number of seeds is set with
reference to $K$. \ddag~ASM fits spherical Gaussian components on each client;
the global cluster model is that of the second-phase aggregator.}
\label{tab:related}
\small
\setlength{\tabcolsep}{5pt}
\begin{tabular}{lcccl}
\toprule
Method & Federated & Unknown $K$ & Unknown $K_g$ & Cluster model \\
\midrule
$k$-means / GMM              & \xmark & \xmark & --          & spherical / Gaussian \\
X-means \citep{pelleg2000xmeans}   & \xmark & \cmark & --     & spherical Gaussian \\
G-means \citep{hamerly2003gmeans}  & \xmark & \cmark & --     & Gaussian \\
DP-GMM \citep{antoniak1974dp}      & \xmark & \cmark & --     & Gaussian \\
\midrule
k-FED \citep{dennis2021kfed}        & \cmark & \xmark & \xmark & spherical \\
FedKmeans \citep{garst2024fedkmeans} & \cmark & \xmark & \xmark & spherical \\
FFCM \citep{stallmann2022ffcm}      & \cmark & \xmark & \xmark & fuzzy / spherical \\
\midrule
SPAHM \citep{yurochkin2019spahm}    & \cmark & \cmark & \xmark & model-agnostic \\
AFCL \citep{zhang2025afcl}          & \cmark & \cmark & (\cmark)\dag & centroid-based \\
FedGEM \citep{ibrahim2026fedgem}    & \cmark & \cmark & \xmark & isotropic Gaussian \\
\midrule
This work                           & \cmark & \cmark & \cmark & spherical Gaussian\ddag \\
\bottomrule
\end{tabular}
\end{table}



Table~\ref{tab:related} summarizes the methods discussed below by what each
requires as input.

\paragraph{Federated clustering with known $K$.}
k-FED \citep{dennis2021kfed} runs $k$-means on each client and clusters the
pooled centroids into $K$ global clusters in a single round. FedKmeans
\citep{garst2024fedkmeans} performs federated Lloyd iterations under a shared
$K$. The federated fuzzy $c$-means methods FFCM-avg1 and FFCM-avg2
\citep{stallmann2022ffcm} use fuzzy memberships and differ in how local
prototypes are averaged. All of these require $K$ at the server.

\paragraph{Federated clustering with unknown $K$.}
AFCL \citep{zhang2025afcl} initializes each client with more seed points than
clusters and lets redundant seeds merge through interaction with the server, so
that the number of distinct clusters emerges as a by-product; clients may also
communicate with the server asynchronously. FedGEM \citep{ibrahim2026fedgem}
casts federated clustering as federated generalized EM: each client fits a
local mixture with $K_g$ components together with an uncertainty set for each
component, and the server groups components whose sets overlap, estimating $K$
as the number of groups. Because this grouping never uses a cluster count,
FedGEM can serve as the second phase of our framework
(Section~\ref{sec:experiments2}). SPAHM \citep{yurochkin2019spahm} takes a related
approach for models trained independently on separate datasets, matching their
parameters under a Bayesian nonparametric prior so that the number of global
components is inferred. These methods relax the assumption of a known $K$ but
retain other cardinality inputs: FedGEM takes each $K_g$ as given, SPAHM
aggregates local models whose sizes were fixed beforehand, and the published
experiments of AFCL draw the initial number of seeds from $[K, 2K]$.

\paragraph{Local cardinality.}
To our knowledge, no federated clustering method estimates both the local
counts $K_g$ and the global count $K$ from data without cardinality inputs. Our
framework treats both as unknown and supplies the local counts that aggregators
such as FedGEM require.

\paragraph{Choosing the number of clusters in centralized clustering.}
X-means \citep{pelleg2000xmeans} splits a centroid when the split improves the
Bayesian Information Criterion, and G-means \citep{hamerly2003gmeans} splits
when a one-dimensional projection of the assigned points fails a normality
test, initializing the children along the principal direction. Split-and-merge
EM \citep{ueda2000smem} uses split and merge moves to escape local optima at a
fixed number of components, whereas \citet{figueiredo2002unsupervised} select
the number of components by annihilating weak components under a
minimum-message-length criterion. Dirichlet-process mixtures
\citep{antoniak1974dp, blei2006variational} infer the number of components
under a nonparametric prior. ASM shares the BIC-driven recursive splitting of X-means and the principal-direction initialization of G-means, but fits each candidate split by EM under a spherical Gaussian model. It also introduces a merge stage that operates on the same components, selection of all hyperparameters on held-out client data alone, and an analysis of when the BIC split criterion fragments a single cluster (Section~\ref{sec:method}), which applies to X-means-type estimators generally. We compare against DP-GMM, X-means and G-means as first-phase estimators in
Section~\ref{sec:experiments2}.

\paragraph{Clustered federated learning.}
Clustered federated learning \citep{ghosh2020ifca, sattler2020cfl} groups
clients with similar data distributions in order to train a separate model for
each group. It clusters clients rather than data points and is complementary
to the problem studied here.

\section{Problem setting}
\label{sec:setup}

A server coordinates $G$ clients. Client $g \in [G]$ holds
$\mathcal{X}_g = \{\vx_{gi}\}_{i=1}^{N_g} \subset \mathbb{R}^d$ and may share model parameters or summary statistics with the server, but never observations. Clients cannot tune hyperparameters jointly, either with one another or with the server: any hyperparameter used on client $g$ must be selected from $\mathcal{X}_g$ alone. Client-server communication is limited to parameters and summary statistics, and the local phase requires no iteration with the server. The pooled data $\mathcal{X} = \bigcup_g \mathcal{X}_g$ belong to $K$ global clusters indexed by $[K]$. Client $g$ holds points from a subset $S_g \subseteq [K]$ of local cardinality $K_g = |S_g|$, where
\begin{equation}
  \bigcup_{g=1}^{G} S_g = [K], \qquad
  2 \le K_g < K \;\text{ for all } g, \qquad
  S_g \cap S_{g'} \neq \emptyset \;\text{ for some } g \neq g'.
  \label{eq:setup}
\end{equation}
Clients differ both in which clusters they hold and in the proportions of those clusters, so the local data distributions are non-IID. Neither the server nor the clients know $K$, $\{K_g\}_{g=1}^{G}$ or $\{S_g\}_{g=1}^{G}$. The goal is to partition $\mathcal{X}$ into global clusters and to estimate $K$ without moving data between clients. 

The framework has two phases. First, each client independently computes an estimate $\hat K_g = \mathcal{A}(\mathcal{X}_g)$ of its local cardinality. Second, an aggregator that requires local cardinalities, such as FedGEM \citep{ibrahim2026fedgem}, takes $\{\hat K_g\}$ in place of the true values and, using client-side computations, returns global cluster parameters and $\hat K$. We make no generative assumption about the data: ASM uses Gaussian components as a working model, and the analysis of Section~\ref{sec:method} is carried out for Gaussian clusters to characterize its behavior, but neither the method nor the experiments require the data to follow a Gaussian mixture. In many of the experiments, clusters are the classes of benchmark datasets. Class labels are used only to construct the client subsets $S_g$ and to evaluate the results, never by the methods.

\section{Adaptive split--merge (ASM) algorithm and Analysis}
\label{sec:method}
\subsection{Algorithm and Analysis} 
ASM starts from a single component containing all of the client's points and
grows the model by recursive binary splitting. To decide whether to split a
component, it computes the eigendecomposition of its covariance matrix: the
leading eigenvector $v_1$, with eigenvalue $\lambda_1$, gives the direction of
greatest spread. Two candidate centroids are formed by perturbing the
component mean along this direction in opposite senses,
$\mu_{\pm} = \mu \pm \delta\sqrt{\lambda_1}\,v_1$, where $\delta$ is the split
factor, and are used to initialize an EM run restricted to the component's
points and capped at 15 iterations. The split is accepted if it improves the
Bayesian Information Criterion (BIC) over the unsplit component
(Appendix~\ref{app:bic}) and each child retains an effective mass, the sum of
its responsibilities, of at least $\tau_{\mathrm{thresh}}$. The points of an
accepted split are reassigned to the child with the highest responsibility,
and the children are tested for further splitting in the next pass. Once no
split is accepted, components are merged greedily, one pair at a time,
whenever their centroids are closer than a multiple of their combined spread,
\begin{equation}
  \lVert \mu_i - \mu_j \rVert_2 \;\le\; \alpha\,(\sigma_i + \sigma_j),
  \label{eq:merge}
\end{equation}
where $\sigma_k = \sqrt{\operatorname{tr}(\Sigma_k)/d}$ is the root-mean-square
per-dimension deviation of component $k$ and $\alpha > 0$ is the merge
factor. Under the spherical model, a component has standard deviation
$\sigma_k$ along every direction, so \eqref{eq:merge} tests whether two
components overlap along the line joining their means, independently of the
dimension; the criterion is of the Davies--Bouldin form
\citep{davies1979cluster}. The pooled mean and spread are recomputed after
each merge, and the estimated local count $K_g$ is the number of components
that remain. Pseudocode is given in Appendix~\ref{ASMpseudo}.

\paragraph{When does ASM split?}
Under simplifying assumptions, the split rule admits a closed form that explains ASM's operating envelope. For a single Gaussian cluster of $n$ points cut at its median along the
leading eigenvector, the split criterion
(\eqref{eq:app-bic} in Appendix~\ref{app:bic}) is met when:
\begin{equation}
  \frac{n}{\log n} \;>\; \frac{d+2}{g(r,d)},
  \qquad
  g(r,d) = -d\log\!\Big(1 - \frac{2r}{\pi d}\Big) \approx \frac{2r}{\pi},
  \qquad
  r = \frac{\lambda_1}{\operatorname{tr}\Sigma/d}.
  \label{eq:threshold}
\end{equation}
The rule is invariant to feature scale and defines a critical cluster size
that falls with the anisotropy $r$ and rises with the dimension $d$. Three consequences follow. Isotropic clusters are protected up to a size that grows with $d$, which accounts for near-exact recovery of isotropic clusters in high dimension. Anisotropy is the binding limitation: an elongated cluster reaches its critical size at roughly $1/r$ of the isotropic value and is then split even though it is a single true cluster. And because every cluster has a finite critical size, larger clients split more readily which makes the merge stage in the algorithm essential. 

\begin{figure}[t]
  \centering
  \includegraphics[width=\linewidth]{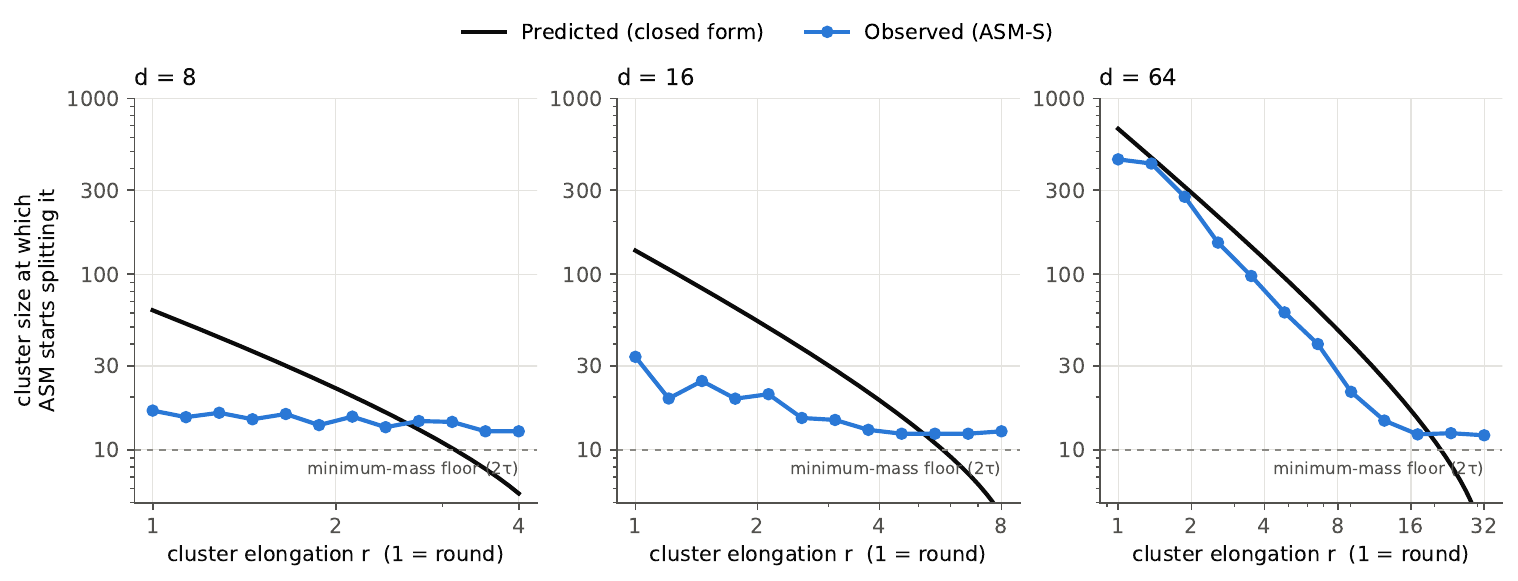}
  \caption{Size at which ASM-S begins to split a single Gaussian cluster, as a
  function of its anisotropy $r$ ($r = 1$ is round). Black: prediction of
  \eqref{eq:threshold} with the population anisotropy. Blue: size above which at
  least half of the proposed splits are accepted (50 draws per point,
  $\delta = 1$, $\tau_{\mathrm{thresh}} = 5$). Below a curve the cluster is kept
  whole; above it, it is split. Dashed: $2\tau_{\mathrm{thresh}}$, the smallest
  size the minimum-mass constraint allows to be split.}
  \label{fig:split-onset}
\end{figure}

\paragraph{Validating the split analysis.}
To test the analysis directly, we apply the split test of ASM-S (ASM with Spherical Gaussians) to a single Gaussian cluster, for which every accepted split is spurious. The cluster has anisotropy $r$, set independently of the dimension, and for each $(d, r)$ we record the size above which at least half of the proposed splits are accepted (Appendix~\ref{app:synthetic-split}). Figure~\ref{fig:split-onset} compares this onset with the prediction of \eqref{eq:threshold}. At $d = 64$ the prediction
tracks the observed onset across an order of magnitude in both $r$ and $n$: a
round cluster is kept whole up to about 450 points, whereas one with $r \approx 9$
is split from about 20. At $d = 8$ and $16$ the prediction is optimistic: ASM-S
splits even round clusters from 15 to 35 points, and the onset sits just above
the smallest size the minimum-mass constraint permits. Both effects have a single
cause. When $n$ is not much larger than $d$, the anisotropy estimated from a
cluster's own points exceeds its population value, and with $r$ measured on the
sample the closed form reproduces the idealized split decision in 94--97\% of
draws. Once a split is accepted, the recursion continues well below the onset, because halving a cluster inflates the sample anisotropy of its children. In low dimension a single cluster of 500 points yields about 50 components before merging at $d = 8$ and about 20 at $d = 16$, approaching the bound set by the minimum-mass constraint (Figure~\ref{fig:premerge}). In this regime the merge stage determines the final count.

\subsection{ASM on pooled benchmark data}
\label{sec:pooled}

The analysis in the previous subsection predicts that ASM is least accurate when a single client holds a large dataset. We therefore evaluate it first in the pooled setting on eight benchmark datasets: four tabular (Waveform, Abalone, FrogA, FrogB) and four image datasets (MNIST, FMNIST, EMNIST, CIFAR-10). Hyperparameters of ASM-S and DP-GMM are selected without labels by the silhouette objective, and results are reported as held-out ARI over ten seeds. As reference points we include X-means, G-means and $k$-means given the true $K^*$ (Table~\ref{tab:fv_pooled_benchmark}).

\begin{table}[h]
\centering
\footnotesize
\caption{Pooled (single-client) benchmark: held-out ARI $\pm$ 95\% CI half-width ($t$, $n = 10$ seeds 42--51); second line $\hat K$ = mean estimated $K$. ASM-S and DP-GMM chosen by the label-free silhouette objective; X-/G-Means have none. Oracle: $k$-means with the true $K$ given (k-means++ initialisation, 10 restarts).The other methods do not use the true $K$.}
\label{tab:fv_pooled_benchmark}
\resizebox{\ifdim\width>\linewidth\linewidth\else\width\fi}{!}{%
\begin{tabular}{lcccccc}
\toprule
Dataset & $K^\ast$ & ASM-S & DP-GMM & X-Means & G-Means & $k$-means, $K$ known \\
\midrule
Waveform & 3 & \makecell{0.256 $\pm$ 0.006 \\ {}$\hat K$ 3.0} & \makecell{0.273 $\pm$ 0.005 \\ {}$\hat K$ 6.3} & \makecell{0.308 $\pm$ 0.016 \\ {}$\hat K$ 6.2} & \makecell{0.208 $\pm$ 0.073 \\ {}$\hat K$ 46.2} & \makecell{0.254 $\pm$ 0.002 \\ {}$K$ = 3} \\
\midrule
Abalone & 8 & \makecell{0.087 $\pm$ 0.006 \\ {}$\hat K$ 7.5} & \makecell{0.084 $\pm$ 0.008 \\ {}$\hat K$ 13.5} & \makecell{0.058 $\pm$ 0.003 \\ {}$\hat K$ 28.3} & \makecell{0.023 $\pm$ 0.003 \\ {}$\hat K$ 129.1} & \makecell{0.086 $\pm$ 0.004 \\ {}$K$ = 8} \\
\midrule
FrogA & 10 & \makecell{0.273 $\pm$ 0.013 \\ {}$\hat K$ 10.6} & \makecell{0.090 $\pm$ 0.003 \\ {}$\hat K$ 100.3} & \makecell{0.124 $\pm$ 0.005 \\ {}$\hat K$ 77.6} & \makecell{0.070 $\pm$ 0.006 \\ {}$\hat K$ 200.0} & \makecell{0.452 $\pm$ 0.034 \\ {}$K$ = 10} \\
\midrule
FrogB & 8 & \makecell{0.153 $\pm$ 0.016 \\ {}$\hat K$ 10.8} & \makecell{0.054 $\pm$ 0.002 \\ {}$\hat K$ 99.8} & \makecell{0.078 $\pm$ 0.004 \\ {}$\hat K$ 76.9} & \makecell{0.046 $\pm$ 0.002 \\ {}$\hat K$ 200.0} & \makecell{0.369 $\pm$ 0.006 \\ {}$K$ = 8} \\
\midrule
MNIST & 10 & \makecell{0.162 $\pm$ 0.010 \\ {}$\hat K$ 81.0} & \makecell{0.224 $\pm$ 0.019 \\ {}$\hat K$ 19.5} & \makecell{0.112 $\pm$ 0.006 \\ {}$\hat K$ 2.0} & \makecell{0.078 $\pm$ 0.000 \\ {}$\hat K$ 200.0} & \makecell{0.643 $\pm$ 0.003 \\ {}$K$ = 10} \\
\midrule
FMNIST & 10 & \makecell{0.070 $\pm$ 0.001 \\ {}$\hat K$ 152.5} & \makecell{0.049 $\pm$ 0.027 \\ {}$\hat K$ 2.5} & \makecell{0.060 $\pm$ 0.001 \\ {}$\hat K$ 200.0} & \makecell{0.062 $\pm$ 0.002 \\ {}$\hat K$ 200.0} & \makecell{0.431 $\pm$ 0.002 \\ {}$K$ = 10} \\
\midrule
EMNIST & 47 & \makecell{0.180 $\pm$ 0.003 \\ {}$\hat K$ 193.2} & \makecell{0.202 $\pm$ 0.001 \\ {}$\hat K$ 198.1} & \makecell{0.017 $\pm$ 0.001 \\ {}$\hat K$ 2.0} & \makecell{0.200 $\pm$ 0.001 \\ {}$\hat K$ 200.0} & \makecell{0.326 $\pm$ 0.002 \\ {}$K$ = 47} \\
\midrule
CIFAR-10 & 10 & \makecell{0.156 $\pm$ 0.009 \\ {}$\hat K$ 87.7} & \makecell{0.259 $\pm$ 0.012 \\ {}$\hat K$ 71.3} & \makecell{0.083 $\pm$ 0.001 \\ {}$\hat K$ 200.0} & \makecell{0.084 $\pm$ 0.002 \\ {}$\hat K$ 200.0} & \makecell{0.647 $\pm$ 0.002 \\ {}$K$ = 10} \\
\bottomrule
\end{tabular}}
\end{table}

\begin{table}[t]
\centering
\caption{Class anisotropy, onset of merging, and the effect of sample size on
ASM-S in the pooled setting. $r_{\mathrm{med}}$: median over classes of
$\lambda_1(\Sigma_k)/(\operatorname{tr}\Sigma_k/d)$, computed from the labels
for analysis only. $0.8\sqrt{r_{\mathrm{med}}}$: predicted merge factor at
which fragments of a typical class are reunited (Appendix~\ref{app:bic}).
$\alpha_{1/2}$: interval of grid values between which the estimated count
first falls to half its value without merging (Table~\ref{tab:fv_merge_alpha_sweep_pooled}).
$\hat K$ and ARI: ASM-S on the full data $\to$ a fixed random 25\% of each
image dataset, with hyperparameters re-selected by the silhouette objective on
the subsample; mean over seeds 42--51. Oracle: $k$-means with $K^*$ given.}
\label{tab:mechanisms}
\small
\setlength{\tabcolsep}{4pt}
\begin{tabular}{lrrrccccc}
\toprule
 & & & & & & \multicolumn{2}{c}{ASM-S} & Oracle \\
\cmidrule(lr){7-8}
Dataset & $d$ & $K^*$ & $r_{\mathrm{med}}$ & $0.8\sqrt{r_{\mathrm{med}}}$
  & $\alpha_{1/2}$ & $\hat K$ & ARI & ARI \\
\midrule
Waveform & 21 & 3  & 6.39  & 2.02 & 1.0--1.25 & 3.0  & 0.256 & 0.254 \\
Abalone  & 7  & 8  & 6.34  & 2.01 & 1.5--2.0  & 7.5  & 0.087 & 0.086 \\
FrogA    & 21 & 10 & 9.74  & 2.50 & 2.0--3.0  & 10.6 & 0.273 & 0.452 \\
FrogB    & 21 & 8  & 10.11 & 2.54 & 2.0--3.0  & 10.8 & 0.153 & 0.369 \\
\midrule
MNIST    & 10 & 10 & 2.07  & 1.15 & 1.0--1.25 & $81.0 \to 26.5$  & $0.162 \to 0.331$ & $0.643 \to 0.638$ \\
FMNIST   & 64 & 10 & 21.12 & 3.68 & $> 3.0$   & $152.5 \to 38.4$ & $0.070 \to 0.203$ & $0.431 \to 0.429$ \\
EMNIST   & 16 & 47 & 4.07  & 1.61 & 1.25--1.5 & $193.2 \to 45.9$ & $0.180 \to 0.222$ & $0.326 \to 0.315$ \\
CIFAR-10 & 64 & 10 & 8.16  & 2.29 & 2.0--3.0  & $87.7 \to 17.3$  & $0.156 \to 0.425$ & $0.647 \to 0.643$ \\
\bottomrule
\end{tabular}
\end{table}

\paragraph{Comparison with other estimators.}
On the tabular datasets, ASM-S is the only method whose estimate is close to $K^*$ throughout ($\hat K = 3.0, 7.5, 10.6, 10.8$). X-means and G-means miss by an order of magnitude or reach the cap of 200 components, and DP-GMM over-estimates on the Frog datasets by a factor of ten. On Waveform and Abalone, the ARI of ASM-S equals that of $k$-means with $K^*$ given (0.256 vs.\ 0.254 and 0.087 vs.\ 0.086), so ARI there is limited by the class structure. On the Frog datasets, ASM-S recovers the count but not the partition, reaching about half the ARI of $k$-means with $K^*$ given. The image datasets behave differently. $k$-means with $K^*$ given reaches ARI of 0.43--0.65, so the representations separate the classes well, yet ASM-S over-estimates $K$ by a factor of 4 to 15, and DP-GMM achieves higher ARI on three of the four. The rest of this section uses the analysis to explain this gap.

\paragraph{Sample size.}
The number of components produced by the split stage grows with the number of points per cluster (Figure~\ref{fig:premerge}). Reducing each image dataset to a fixed random 25\% (Table~\ref{tab:mechanisms}) lowers the estimate of ASM-S by a factor of 3 to 5 and raises its ARI on all four datasets, while $k$-means with $K^*$ given is unchanged. The subsample is therefore not an easier clustering problem and the improvement comes from the smaller sample. Because the hyperparameters are re-selected on the subsample, the comparison reflects both the smaller sample and the re-selected configuration. 

\paragraph{The merge stage.}
Table~\ref{tab:fv_merge_alpha_sweep_pooled} in the Appendix holds the splits fixed and varies the merge factor. Without merging, the split stage over-estimates $K$ on every dataset, by a factor of 1.5 to 15, as expected for pooled sample sizes. The merge rule does not act for $\alpha \le 0.75$ on any dataset. It removes most of the excess as it starts acting. For instance, on Waveform it reduces 28.8 components to the three classes and triples ARI. The analysis estimates that merging begins for fragments of a class with anisotropy $r$ only when $\alpha \gtrsim 0.8\sqrt{r}$ (Appendix~\ref{app:bic}). Evaluated at the median class anisotropy of each dataset (Table~\ref{tab:mechanisms}), this bound matches, to within one grid step, the value of $\alpha$ at which the estimated count first halves on seven of the eight datasets. The exception is Waveform, whose merging begins earlier. The bound also accounts for FMNIST. Its median class anisotropy is 21, so merging requires $\alpha \approx 3.7$, beyond every value tested, and the estimate stays at the split-stage count. On MNIST and EMNIST, the bound is met at moderate $\alpha$,
but larger values begin to join distinct classes, and the estimate passes from tens of components to about one between adjacent grid values, so no single $\alpha$ recovers $K^*$. The silhouette objective also leaves some ARI unused on these datasets (0.200 at $\alpha = 1.25$ against 0.162 selected on
MNIST).

\paragraph{Summary.}
The pooled benchmarks confirm the role of sample size identified by the analysis. The anisotropy bound predicts where merging begins, but not whether it can stop at the right count. On the tabular datasets a single merge factor brings the estimate close to $K^*$; on the image datasets none does. Merging is absent within the searched range on FMNIST, whose classes are the most anisotropic, and on MNIST and EMNIST the estimate passes from tens of components to about one between adjacent values of $\alpha$. On these data the single-parameter merge rule may be the main limitation of ASM-S.

\section{ASM in the Federated Setting}
\label{sec:experiments2}

We evaluate the two-phase pipeline end to end. In Phase~1, each client estimates its local cardinality $K_g$ from its own data; in Phase~2, FedGEM receives the estimates $\{\hat K_g\}$ in place of the true values and returns the global cluster structure and $\hat K$. We compare ASM-S with three other
label-free Phase~1 methods, DP-GMM, X-means and G-means, and with an oracle that supplies the true $K_g$. Every client selects its hyperparameters independently, by the silhouette of its Phase~1 clustering on its own held-out points, and passes only $\hat K_g$ to FedGEM.

\paragraph{Client partitions.}
For each dataset we use $G$ clients (Table \ref{tab:datasets}), and each client is assigned a random subset of the classes, $0.55K$ on average, subject to \eqref{eq:setup}. Points
of a class are disjoint across the clients that hold it. We construct two
partitions. In partition~A, each class is divided equally among the clients
that hold it. Since each class is held by about $0.55K$ clients, a client holds
on average about $1/(0.55K)$ of a class's points: about 18\% for $K = 10$ and
4\% for EMNIST, below the 25\% subsample of Section~\ref{sec:pooled} on every
image dataset. Under partition~A, however, client size grows almost in
proportion to $K_g$, which could let a Phase~1 method infer $K_g$ from client
size rather than from cluster structure. In partition~B, each client keeps the
same classes, but its size is drawn independently of $K_g$ from
$[0.5N_{\mathrm{med}}, 2N_{\mathrm{med}}]$, where $N_{\mathrm{med}}$ is the
median client size under partition~A. On the image datasets, the elasticity of
client size with respect to $K_g$ is about one under partition~A and about
zero under partition~B (Appendix~\ref{app:points-per-cluster}); construction
details are given in Appendix~\ref{app:partitions}.

\paragraph{Results under partition A.}
Table~\ref{tab:fv_federated_realistic} reports downstream ARI and the error of
the global estimate. ASM-S attains the highest mean ARI among the label-free
methods on four of the eight datasets (Waveform, FrogA, FrogB and MNIST) and
trails the best label-free method by at most 0.024 on the other four. Averaged
over the eight datasets, its ARI is 0.333, against 0.256 for X-means, 0.214
for DP-GMM, 0.163 for G-means and 0.361 for the oracle that supplies the true
$K_g$. ASM-S is within 0.03 of the oracle on five datasets and exceeds it on
FrogA. Its global estimates are also the most reliable: $|\hat K - K^*|$ never
exceeds 21 for ASM-S and averages 6.9, compared with 5.4 for the oracle,
whereas it reaches 188 for DP-GMM and 263 for G-means. DP-GMM is competitive in
ARI on Waveform, Abalone and EMNIST, but its global estimate is off by 44.3 on
FrogA, 28.9 on FrogB, 101.1 on FMNIST and 187.8 on CIFAR-10. The image
datasets on which ASM-S over-fragmented in the pooled setting behave
differently here: on MNIST, FMNIST and CIFAR-10 its ARI rises from 0.162,
0.070 and 0.156 in the pooled setting to 0.418, 0.250 and 0.380, consistent
with the smaller per-client cluster sizes.

\begin{table}[h]
\centering
\small
\caption{Full federated pipeline, Regime A, per-client hyperparameter selection. Global ARI $\pm$ 95\% CI half-width (t, n = 10 seeds 42--51); second line $\hat K$ = mean estimated global K, $|\Delta K|$ = mean $\hat K$ $-$ $K^\ast$. Mean row = unweighted mean over the 8 datasets. CIFAR-10 G-Means: 5 timed-out seeds filled from an identical rerun.}
\label{tab:fv_federated_realistic}
\resizebox{\textwidth}{!}{%
\begin{tabular}{lccccc}
\toprule
Dataset & ASM-S & DP-GMM & X-Means & G-Means & Oracle-K \\
\midrule
Waveform & \makecell{0.258 $\pm$ 0.011 \\ $\hat K$ 4.5 $\cdot$ $|\Delta K|$ 1.5} & \makecell{0.256 $\pm$ 0.027 \\ $\hat K$ 4.2 $\cdot$ $|\Delta K|$ 1.2} & \makecell{0.249 $\pm$ 0.027 \\ $\hat K$ 5.4 $\cdot$ $|\Delta K|$ 2.4} & \makecell{0.111 $\pm$ 0.023 \\ $\hat K$ 29.9 $\cdot$ $|\Delta K|$ 26.9} & \makecell{0.339 $\pm$ 0.024 \\ $\hat K$ 2.0 $\cdot$ $|\Delta K|$ 1.0} \\
\midrule
Abalone & \makecell{0.089 $\pm$ 0.015 \\ $\hat K$ 4.9 $\cdot$ $|\Delta K|$ 3.1} & \makecell{0.100 $\pm$ 0.005 \\ $\hat K$ 8.3 $\cdot$ $|\Delta K|$ 2.9} & \makecell{0.098 $\pm$ 0.007 \\ $\hat K$ 16.7 $\cdot$ $|\Delta K|$ 8.7} & \makecell{0.080 $\pm$ 0.012 \\ $\hat K$ 24.7 $\cdot$ $|\Delta K|$ 16.7} & \makecell{0.103 $\pm$ 0.007 \\ $\hat K$ 6.0 $\cdot$ $|\Delta K|$ 2.0} \\
\midrule
FrogA & \makecell{0.634 $\pm$ 0.118 \\ $\hat K$ 7.9 $\cdot$ $|\Delta K|$ 3.7} & \makecell{0.243 $\pm$ 0.088 \\ $\hat K$ 54.3 $\cdot$ $|\Delta K|$ 44.3} & \makecell{0.381 $\pm$ 0.104 \\ $\hat K$ 30.6 $\cdot$ $|\Delta K|$ 20.6} & \makecell{0.174 $\pm$ 0.034 \\ $\hat K$ 131.2 $\cdot$ $|\Delta K|$ 121.2} & \makecell{0.601 $\pm$ 0.104 \\ $\hat K$ 7.9 $\cdot$ $|\Delta K|$ 2.1} \\
\midrule
FrogB & \makecell{0.401 $\pm$ 0.097 \\ $\hat K$ 8.8 $\cdot$ $|\Delta K|$ 3.4} & \makecell{0.234 $\pm$ 0.076 \\ $\hat K$ 36.9 $\cdot$ $|\Delta K|$ 28.9} & \makecell{0.252 $\pm$ 0.059 \\ $\hat K$ 29.2 $\cdot$ $|\Delta K|$ 21.2} & \makecell{0.070 $\pm$ 0.014 \\ $\hat K$ 165.6 $\cdot$ $|\Delta K|$ 157.6} & \makecell{0.451 $\pm$ 0.093 \\ $\hat K$ 5.9 $\cdot$ $|\Delta K|$ 2.1} \\
\midrule
MNIST & \makecell{0.418 $\pm$ 0.036 \\ $\hat K$ 18.5 $\cdot$ $|\Delta K|$ 8.5} & \makecell{0.355 $\pm$ 0.041 \\ $\hat K$ 8.1 $\cdot$ $|\Delta K|$ 1.9} & \makecell{0.198 $\pm$ 0.044 \\ $\hat K$ 3.6 $\cdot$ $|\Delta K|$ 6.4} & \makecell{0.287 $\pm$ 0.039 \\ $\hat K$ 43.8 $\cdot$ $|\Delta K|$ 33.8} & \makecell{0.446 $\pm$ 0.048 \\ $\hat K$ 9.4 $\cdot$ $|\Delta K|$ 1.0} \\
\midrule
FMNIST & \makecell{0.250 $\pm$ 0.029 \\ $\hat K$ 16.6 $\cdot$ $|\Delta K|$ 7.0} & \makecell{0.178 $\pm$ 0.044 \\ $\hat K$ 111.1 $\cdot$ $|\Delta K|$ 101.1} & \makecell{0.274 $\pm$ 0.026 \\ $\hat K$ 21.1 $\cdot$ $|\Delta K|$ 11.1} & \makecell{0.249 $\pm$ 0.023 \\ $\hat K$ 38.3 $\cdot$ $|\Delta K|$ 28.3} & \makecell{0.252 $\pm$ 0.027 \\ $\hat K$ 8.0 $\cdot$ $|\Delta K|$ 2.0} \\
\midrule
EMNIST & \makecell{0.237 $\pm$ 0.006 \\ $\hat K$ 48.3 $\cdot$ $|\Delta K|$ 7.3} & \makecell{0.250 $\pm$ 0.008 \\ $\hat K$ 67.8 $\cdot$ $|\Delta K|$ 20.8} & \makecell{0.193 $\pm$ 0.022 \\ $\hat K$ 25.7 $\cdot$ $|\Delta K|$ 21.3} & \makecell{0.164 $\pm$ 0.009 \\ $\hat K$ 309.9 $\cdot$ $|\Delta K|$ 262.9} & \makecell{0.257 $\pm$ 0.011 \\ $\hat K$ 49.5 $\cdot$ $|\Delta K|$ 2.5} \\
\midrule
CIFAR-10 & \makecell{0.380 $\pm$ 0.029 \\ $\hat K$ 30.9 $\cdot$ $|\Delta K|$ 20.9} & \makecell{0.097 $\pm$ 0.020 \\ $\hat K$ 197.8 $\cdot$ $|\Delta K|$ 187.8} & \makecell{0.404 $\pm$ 0.058 \\ $\hat K$ 18.2 $\cdot$ $|\Delta K|$ 8.2} & \makecell{0.169 $\pm$ 0.016 \\ $\hat K$ 97.2 $\cdot$ $|\Delta K|$ 87.2} & \makecell{0.442 $\pm$ 0.017 \\ $\hat K$ 40.8 $\cdot$ $|\Delta K|$ 30.8} \\
\bottomrule
\end{tabular}}
\end{table}

\paragraph{Results under partition B.}
Partition~B tests whether a Phase~1 method recovers $K_g$ from cluster
structure or merely from client size. The test matters most for ASM, whose
split criterion depends on sample size \eqref{eq:threshold}. The estimates of
ASM-S are unaffected by the change of partition: pooled over datasets and
seeds, the relative error of its global estimate $\hat K$ is statistically
equivalent across the two partitions within a margin of $\pm 0.2$ (two
one-sided tests), and no individual dataset shows a significant difference.
X-means and G-means, by contrast, change significantly on three and four of
the eight datasets, respectively (Appendix~\ref{app:partition-robustness}).
The accuracy of ASM-S under partition~A therefore does not rely on the
correlation between client size and $K_g$. The true and estimated local
cardinalities of every client, for all methods and both partitions, are shown
in Figures~\ref{fig:localk-hist-all} and~\ref{fig:localk-hist}.

\paragraph{Computational cost.}
Table~\ref{tab:runtime} reports the mean Phase~1 wall-clock time per client.
ASM-S is the fastest method on six of the eight datasets and has the lowest
mean, 339\,ms against 558\,ms for G-means, 1{,}110\,ms for DP-GMM and
1{,}943\,ms for X-means; G-means is faster on MNIST and FMNIST. Timings
reflect the implementations used (Appendix~\ref{app:details}) and were
measured on CPU-only Google Cloud virtual machines running Debian~12, using Python~3.11, scikit-learn~1.6, and NumPy~2.0. No GPUs were used. Runtime measurements were performed on an N2 instance with 32~vCPUs (Intel Xeon), while other experiments used an E2 instance with 12~vCPUs (AMD EPYC~7B12, 12\,GB RAM).

\begin{table}[htbp]
\centering
\caption{Mean Phase-1 wall-clock time per client (ms) in the federated benchmark. Bold: fastest.}
\label{tab:runtime}
\small
\begin{tabular}{lrrrr}
\toprule
Dataset & \asmsph & DP-GMM & X-Means & G-Means \\
\midrule
Waveform & \textbf{266} & 309 & 865 & 500 \\
Abalone & \textbf{193} & 529 & 1,654 & 490 \\
FrogA & \textbf{410} & 1,041 & 3,217 & 679 \\
FrogB & \textbf{314} & 1,072 & 2,883 & 710 \\
MNIST & 96 & 399 & 565 & \textbf{70} \\
FMNIST & 211 & 528 & 3,478 & \textbf{114} \\
EMNIST & \textbf{1,135} & 4,839 & 2,643 & 1,763 \\
CIFAR-10 & \textbf{90} & 166 & 238 & 136 \\
\midrule
\textit{Mean} & \textbf{339} & 1,110 & 1,943 & 558 \\
\bottomrule
\end{tabular}
\end{table}

\section{Conclusion}
\label{sec:conclusion}

We studied federated clustering when neither the local nor the global number of
clusters is known, a setting that arises when independent sites, such as the
clients of an equipment manufacturer, know no more about their cluster
structure than the server does. We proposed a two-phase framework in which each
client estimates its local count independently and a federated aggregator
reconciles the results, and introduced ASM as a label-free first phase that
combines BIC-driven splitting with a merge stage.

Our analysis characterizes when such an estimator works. The split criterion
admits a critical cluster size that falls with anisotropy and rises with
dimension, and beyond it a single cluster is fragmented until the minimum-mass
constraint binds, so the merge stage, rather than the split criterion,
determines the final count. On pooled benchmarks, over-fragmentation grows with
the number of points per class and shrinks when the sample is reduced, and the
onset of merging follows a bound derived from class anisotropy. In the federated
setting, where each client holds only part of each class, ASM-S with FedGEM
comes within 0.03 ARI of the oracle with true local counts on five of eight
datasets, gives the most reliable global estimates of the label-free methods,
is robust when client size is decoupled from local cardinality, and is the
fastest on average.

Several limitations remain. On the pooled image datasets no single merge factor
recovers the true count: merging either does not begin within the searched
range or passes from tens of components to about one between adjacent values,
and the silhouette objective does not always select the most accurate setting.
A merge criterion that accounts for component shape, or that stops at a
data-driven count, is a natural next step. In deployment, clusters that
dominate a site's data, such as normal operation in the OEM setting, can exceed
the critical size at every site, and their fragmentation must then be corrected
by the merge stage or the aggregator. Our treatment of disclosure is informal,
and we do not provide formal privacy guarantees. Finally, the analysis relies
on idealizations (Gaussian clusters, hard assignments and population moments)
whose effects we quantify empirically rather than bound.

\section*{Generative AI Usage Statement}

We used a large language model assistant (Claude, Anthropic) in the following
ways. \emph{Paper writing:} drafting and editing of text in all sections,
including the introduction, related work, problem setup, experimental
narrative and conclusion; and suggesting keywords and the choice of primary area. \emph{Literature:} suggesting related work (X-means, G-means, split-and-merge EM, parameter-matching aggregation and
clustered federated learning). \emph{Theory:} deriving the closed-form BIC split criterion for the spherical
Gaussian model, its reduction to an anisotropy threshold, the merge-factor
bound $\alpha \gtrsim 0.8\sqrt{r}$, and the termination argument
(Section~\ref{sec:method} and Appendix~\ref{app:bic}), and editing the wording of the experimental sections and appendices. \emph{Methodology
and implementation:} proposing diagnostic experiments (the oracle baseline,
the merge-factor sweep with fixed splits, the subsampling experiment and the
class-level anisotropy diagnostic), and writing the code and figures for the
single-cluster validation experiment (Section~\ref{sec:method} and Appendix~\ref{app:synthetic-split}). \emph{Code:} assisting with the implementation of ASM and the experimental pipeline and running and debugging experiment and analysis scripts on top of the implementation of ASM and FedGEM, \emph{Interpretation:} suggesting explanations of the benchmark results, several of which we tested and revised or discarded. The authors checked all derivations, verified every reported number against the experimental outputs, and take full responsibility for the content of the paper.

\section*{Reproducibility statement}
The ASM algorithm is specified in Section \ref{sec:method}, and complete pseudocode in Appendix \ref{ASMpseudo}. The analysis of our implementation of FedGEM compared to the paper's results is in Appendix \ref{app:repro}. The hyperparameter selection protocol, including the construction of the held-out client split, number of clients and per-client sample sizes, is described in Appendix \ref{app:details} and \ref{app:partitions}; no labels or ground-truth cluster counts are used at any point in selection. The codebase is attached as a zip file in supplementary material. 

\bibliography{references}
\bibliographystyle{iclr2026_conference}

\newpage
\appendix
\section*{Supplementary material}
\addcontentsline{toc}{section}{Supplementary material}



\section{The ASM Algorithm pseudocode}\label{ASMpseudo}
\begin{algorithm}[H]
\caption{Adaptive Split--Merge (ASM)}
\label{alg:adaptive_split}
\begin{algorithmic}[1]

\REQUIRE Local dataset $\mathcal{D}_g$, split factor $\delta$, minimum mass threshold $\tau_{\mathrm{thresh}}$, merge factor $\alpha$, EM iteration cap $T=15$

\STATE $K_g \leftarrow 1$
\STATE Initialize Gaussian using the empirical mean of $\mathcal{D}_g$

\REPEAT
    \STATE SplitAccepted $\leftarrow$ False

    \FORALL{components $k=1,\dots,K_g$}

        \STATE Fit a one-component model to the points of component $k$ by EM ($\le T$ iterations)
        \STATE Compute the largest eigenvalue $\lambda_1$ and corresponding eigenvector $v_1$ of the responsibility-weighted covariance
        \STATE Initialize child means as
        $
        \mu_k^{(1)},\mu_k^{(2)}
        \leftarrow
        \mu_k \pm \delta\sqrt{\lambda_1}\,v_1
        $
        \STATE Fit a two-component model to the same points by EM ($\le T$ iterations), initialized at $\mu_k^{(1)},\mu_k^{(2)}$
        \STATE Compute the BIC improvement $\Delta_{\mathrm{BIC}}$ of the two-component model over the one-component model

        \IF{$\Delta_{\mathrm{BIC}}>0$ \AND\ both children have effective mass $\ge \tau_{\mathrm{thresh}}$}
            \STATE Mark $k$ for splitting; each point of $k$ goes to the child it is most responsible for
            \STATE SplitAccepted $\leftarrow$ True
        \ENDIF

    \ENDFOR
    \STATE Replace every marked parent by its two children \COMMENT{$K_g\leftarrow K_g+\#\text{splits}$}

\UNTIL{SplitAccepted = False}

\REPEAT
    \STATE MergeOccurred $\leftarrow$ False
    \IF{some pair $(i,j)$ satisfies $\|\mu_i-\mu_j\|_2 \le \alpha(\sigma_i+\sigma_j)$, where $\sigma_k=\sqrt{\operatorname{tr}(\Sigma_k)/d}$}
        \STATE Merge the first such pair: size-weighted mean, pooled points, $\sigma$ recomputed \COMMENT{$K_g\leftarrow K_g-1$}
        \STATE MergeOccurred $\leftarrow$ True
    \ENDIF
\UNTIL{MergeOccurred = False}

\RETURN Refined local mixture model

\end{algorithmic}
\end{algorithm}

\section{Analysis of the ASM Split Criterion}
\label{app:bic}

\subsection{The criterion}

ASM accepts a proposed split of component $k$ when it improves a BIC-style
score computed on the $n_k$ points of component $k$ alone,
\begin{equation}
\label{eq:app-bic}
\Delta_{\mathrm{BIC}} \;=\; L_2 - L_1 - \frac{d+2}{2}\log n_k ,
\end{equation}
and the split is accepted when $\Delta_{\mathrm{BIC}} > 0$, that is, when
\begin{equation}
\label{eq:app-accept}
2(L_2 - L_1) \;>\; (d+2)\log n_k .
\end{equation}

\paragraph{Local log-likelihoods.}
$L_M$, $M \in \{1,2\}$, is computed from an $M$-component spherical Gaussian
mixture, $\Sigma_m = s_m^2 I_d$, fitted by EM to the $n_k$ points. The EM
objective decomposes into a mixing-weight term and a component term,
\begin{equation}
Q \;=\; \sum_{i=1}^{n_k}\sum_{m=1}^{M} \gamma_{im}\log w_m
\;+\; \underbrace{\sum_{i=1}^{n_k}\sum_{m=1}^{M} \gamma_{im}
\log \mathcal{N}\!\left(x_i \mid \mu_m, s_m^2 I_d\right)}_{L_M},
\end{equation}
where $\gamma_{im}$ are the responsibilities from the final E-step and $w_m$
the mixing weights. We use the component term as the local log-likelihood
$L_M$. For $M=1$, $\gamma_{i1}=1$ and $\log w_1 = 0$, so $L_1$ is the ordinary
single-Gaussian log-likelihood. For $M=2$, $L_2$ omits the mixing-weight
term, which at the M-step equals $-n_k H(w)$ with $H$ the entropy of the
mixing weights; $L_2$ is therefore not the observed-data log-likelihood
$\sum_i \log\sum_m w_m\,\mathcal{N}(x_i \mid \mu_m, s_m^2 I_d)$, and relative
to the complete-data objective it credits a split with up to $n_k\log 2$.

\paragraph{Penalty.}
Splitting one component into two adds a $d$-dimensional mean, a variance and
a mixing weight, that is, $d+2$ free parameters. If the covariance is instead
fixed at $I_d$, the variance is not estimated and the penalty becomes
$\tfrac{d+1}{2}\log n_k$.

\subsection{Closed form under hard assignment}

Under the spherical model, the maximum-likelihood estimates are
$\hat\mu = \bar x$ and $\hat s^2 = \operatorname{tr}\hat\Sigma/d$, and at the
optimum the quadratic term of the log-likelihood reduces to a constant, giving
\begin{equation}
L_1 = -\frac{n_k d}{2}\log\!\big(2\pi\hat s^2\big) - \frac{n_k d}{2}.
\end{equation}
Assigning each point wholly to one of two components of sizes $\rho n_k$ and
$(1-\rho)n_k$, with variances $\hat s_1^2$ and $\hat s_2^2$, each component
contributes a term of the same form, so
\begin{equation}
L_2 = -\frac{d}{2}\sum_{m=1}^{2} n_m\log\!\big(2\pi\hat s_m^2\big)
      - \frac{n_k d}{2}.
\end{equation}
The constants cancel. Writing
$\tilde s^2 = (\hat s_1^2)^{\rho}(\hat s_2^2)^{1-\rho}$ for the weighted
geometric mean of the component variances,
\begin{equation}
L_2 - L_1 = \frac{n_k d}{2}\log\frac{\hat s^2}{\tilde s^2},
\end{equation}
and \eqref{eq:app-accept} becomes
\begin{equation}
\label{eq:app-criterion}
n_k d \log\frac{\hat s^2}{\tilde s^2} \;>\; (d+2)\log n_k .
\end{equation}
Because the variances enter only through their ratio, the criterion is
invariant under any rescaling $x \mapsto cx$ of the feature space. This does
not hold when the covariance is fixed at $I_d$: the gain is then
$\tfrac12\,\tfrac{n_1 n_2}{n_k}\lVert\hat\mu_1 - \hat\mu_2\rVert^2$, which scales
as $c^2$ while the penalty does not. Had the mixing-weight term been retained
in $L_2$, the left-hand side of \eqref{eq:app-criterion} would carry an
additional $-2n_k H(\rho)$.

\subsection{Reduction to an anisotropy threshold}

Consider a single Gaussian cluster with covariance eigenvalues
$\lambda_1 \ge \dots \ge \lambda_d$, cut at its median along the leading
eigenvector, as ASM's initialization proposes, so that $\rho = \tfrac12$. On
either side of the cut, the coordinate along the split direction is
half-normal with variance $\lambda_1(1 - 2/\pi)$, while the orthogonal
coordinates, being independent of it, are unchanged. In the large-sample
limit, therefore,
\begin{equation}
\hat s^2 = \bar\lambda \equiv \frac{\operatorname{tr}\Sigma}{d},
\qquad
\tilde s^2 = \bar\lambda - \frac{2}{\pi}\frac{\lambda_1}{d}.
\end{equation}
Define the anisotropy ratio $r = \lambda_1/\bar\lambda \in [1, d]$, equal to
one for an isotropic cluster and large for an elongated one. Since
$-\log(1-x) \ge x$, the left-hand side of \eqref{eq:app-criterion} is at
least $2n_k r/\pi$, with near equality when $2r/(\pi d) \ll 1$. The split is
therefore accepted when
\begin{equation}
\label{eq:app-threshold}
\frac{n_k}{\log n_k} \;>\; \frac{\pi(d+2)}{2r}.
\end{equation}

\subsection{Consequences}

\paragraph{Every cluster has a critical size.}
Equation~\eqref{eq:app-threshold} defines, for each anisotropy $r$, a critical
size $n^*(r)$: a single true cluster with fewer points is not split, and one
with more is. Since $r \ge 1$, no cluster larger than about
$2\tau_{\mathrm{thresh}}$ is protected at every size; an isotropic cluster is
split once $n_k/\log n_k$ exceeds $(d+2)/g(1,d)$.

\paragraph{Anisotropy is the binding limitation.}
Since $g(r,d) \approx 2r/\pi$, the critical size decreases roughly as $1/r$, so
an elongated cluster is fragmented at a fraction of the size at which an
isotropic cluster of the same dimension would be.

\paragraph{Dimension protects against splitting.}
The critical size increases with $d$: the penalty grows with the number of
parameters, while the gain per point from a cut along a single direction,
$g(r,d)$, does not grow with $d$ and decreases toward $2r/\pi$. Isotropic
clusters in high dimension are therefore split only at large sample sizes.

\paragraph{More data per client makes splitting more permissive.}
For a fixed cluster shape, the criterion is met once $n_k$ exceeds $n^*(r)$, so
ASM is best matched to the federated regime of moderately sized clients and
over-splits toward the centralized limit, in which one client holds the entire
dataset.

\paragraph{Termination.}
The minimum-mass constraint guarantees termination: a component is split only
if both children retain effective mass at least $\tau_{\mathrm{thresh}}$, so no
component with fewer than about $2\tau_{\mathrm{thresh}}$ points is split and
the local count before merging satisfies $K_g \lesssim n_g/\tau_{\mathrm{thresh}}$.
At the population level the criterion would stop the recursion much earlier.
A median cut halves the number of points, and since each child's leading
eigenvalue is at most $\lambda_1$ while its trace falls by $(2/\pi)\lambda_1$,
its anisotropy is at most $r/\big(1 - 2r/(\pi d)\big)$; for a round cluster this
is only $d/(d - 2/\pi)$. The children would then fail the criterion after a
number of levels logarithmic in $n_k$. In finite samples, however, halving a
component inflates the sample anisotropy of its children, and the recursion
continues until the minimum-mass constraint binds in low dimension
(Appendix~\ref{app:synthetic-split}).

\paragraph{Role of the merge stage.}
Because the split stage protects no cluster larger than about
$2\tau_{\mathrm{thresh}}$, the merge stage prevents the fragmentation of large
clusters. A median split of a Gaussian with anisotropy $r$ and mean
per-dimension variance $\bar\lambda$ produces children whose means are
$2\sqrt{2/\pi}\,\sqrt{\lambda_1} \approx 1.6\sqrt{r\bar\lambda}$ apart, each with
per-dimension deviation close to $\sqrt{\bar\lambda}$ when $r \ll d$. The merge
rule \eqref{eq:merge} therefore reunites them whenever $\alpha \gtrsim 0.8\sqrt{r}$,
where $r$ is the anisotropy of the component that was split, as estimated from
its points. Spurious splits of round clusters are undone for $\alpha \gtrsim 0.8$
at the population level, whereas those of elongated clusters, or of small
fragments whose sample anisotropy is inflated, require a larger $\alpha$, which
in turn risks merging genuinely distinct clusters. Anisotropy thus limits both
stages: it makes spurious splits more likely and makes them harder to undo.
\subsection{Approximations}

The analysis makes three idealizations. First, it assumes hard assignments.
At any parameter values,
$\sum_m \gamma_{im}\log\mathcal{N}_m \le \max_m \log\mathcal{N}_m$, so the
component-term gain obtained by EM does not exceed the hard-assignment gain,
and \eqref{eq:app-threshold} is a lower bound on the size and anisotropy at
which splits are accepted. The gap can be substantial for a spurious split of
a single Gaussian, where responsibilities near the cut are close to
$\tfrac12$ and EM draws the two components together; with EM capped at 15
iterations, the realized gain also depends on the initialization. Second, the
factor $2/\pi$ is exact only for a median cut of a Gaussian parent: the
children of a cut are truncated rather than Gaussian, so the factor is
approximate beyond the first level, and a genuinely bimodal cluster separates
further and is correctly split more readily. Third, the analysis uses population moments. With $n_k$ not much larger than
$d$, the anisotropy estimated from the component's own points exceeds its
population value (Appendix~\ref{app:synthetic-split}), so splits are accepted
at smaller sizes than \eqref{eq:app-threshold} predicts with the population $r$.

\section{Synthetic Validation of the Split Analysis}
\label{app:synthetic-split}

\paragraph{Setup.}
Each dataset is a single Gaussian cluster in $\mathbb{R}^d$ with covariance
$\operatorname{diag}(\kappa, 1, \dots, 1)$ and
$\kappa = r(d-1)/(d-r)$, which gives population anisotropy exactly $r$ in any
dimension and so separates the effects of shape and dimension. We use
$d \in \{8, 16, 64\}$, twelve values of $r$ spaced logarithmically from $1$ to
$d/2$, and eighteen cluster sizes from $6$ to $5000$. Because the data contain one
cluster, any accepted split is spurious. The split test follows
Algorithm~\ref{ASMpseudo}: EM initialized at $\mu \pm \delta\sqrt{\lambda_1}\,v_1$,
at most 15 iterations, the component-term likelihood of
Appendix~\ref{app:bic}, and acceptance when $\Delta_{\mathrm{BIC}} > 0$ and both
children have effective mass at least $\tau_{\mathrm{thresh}} = 5$. For
comparison we also evaluate the idealized split of the analysis: a hard cut
through the mean orthogonal to $v_1$, without a mass constraint. Acceptance
probabilities use 50 independent draws per configuration, and component counts
before merging use 20 draws of the full recursive split stage. The onset in
Figure~\ref{fig:split-onset} is the size above which at least half of the
splits are accepted at every larger size, interpolated logarithmically between
grid points.

\paragraph{Sample anisotropy.}
The closed form \eqref{eq:threshold} is stated in terms of the population
anisotropy, but the split test sees only the cluster's own points. For an
isotropic cluster the leading sample eigenvalue concentrates near
$(1 + \sqrt{d/n})^2$ while the mean eigenvalue stays near one
\citep{johnstone2001}, so the sample anisotropy is approximately
$\hat r \approx (1 + \sqrt{d/n})^2$, well above one when $n$ is comparable to
$d$. Substituting $\hat r$ into the criterion gives a self-consistent onset of
about 250 points for a round cluster at $d = 64$, in agreement with the
idealized hard split (248 points), compared with 671 from the population value.
Across all configurations, the criterion evaluated with the sample anisotropy
predicts the outcome of the idealized split in 97\%, 94\% and 97\% of draws at
$d = 8$, $16$ and $64$, against 90\%, 85\% and 92\% with the population
anisotropy. The analysis is therefore accurate once $r$ is read as the
anisotropy of the points being tested.

\paragraph{Soft assignment and initialization.}
Soft EM raises the onset relative to the idealized hard split for near-isotropic
clusters in high dimension, from 248 to about 450 points for a round cluster at
$d = 64$, consistent with the lower-bound argument of Appendix~\ref{app:bic}.
For elongated clusters the two coincide. The onset is unchanged for
$\delta \in \{0.5, 1, 2\}$ in every configuration, so the split factor is not a
sensitive hyperparameter.

\begin{figure}[t]
  \centering
  \includegraphics[width=0.6\linewidth]{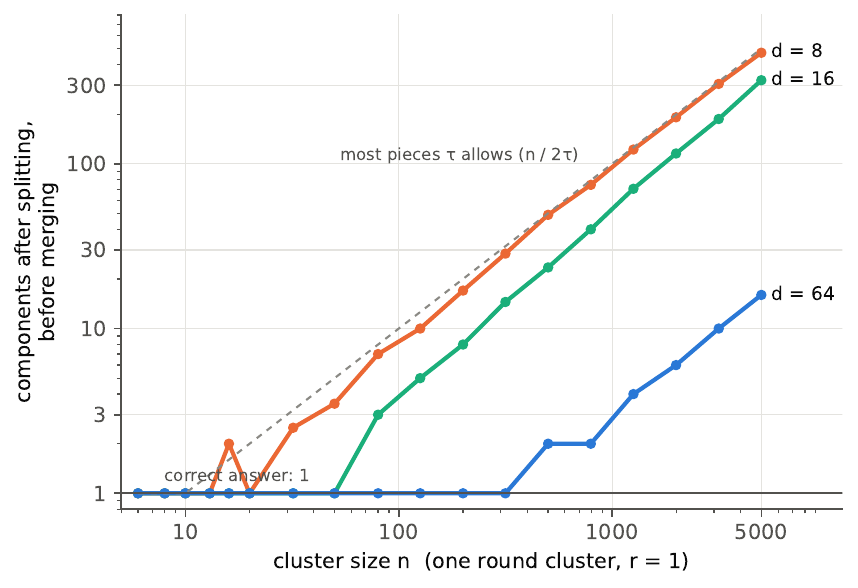}
  \caption{Number of components produced by the split stage of ASM-S, before
  merging, for a single round Gaussian cluster ($r = 1$) of $n$ points; the
  correct count is one. Median over 20 draws, $\delta = 1$,
  $\tau_{\mathrm{thresh}} = 5$. Dashed: the bound $n/(2\tau_{\mathrm{thresh}})$
  set by the minimum-mass constraint.}
  \label{fig:premerge}
\end{figure}

\paragraph{The split stage does not stop at the onset.}
Figure~\ref{fig:premerge} reports the number of components produced by the
split stage, before merging, for a single round cluster. Beyond the onset the
count grows roughly in proportion to $n$. At $d = 8$ it follows the bound
$n / (2\tau_{\mathrm{thresh}})$ imposed by the minimum-mass constraint, reaching
about 470 components for $n = 5000$; at $d = 16$ it reaches about 320, and at
$d = 64$ about 16. The recursion continues because the children of a split are
not rounder than their parent: each accepted split halves the number of points,
which inflates the sample anisotropy of the children, and in a representative
run at $d = 8$ the median sample anisotropy rises from $1.06$ at the root to
about $2.7$ eleven levels down. Termination is therefore governed by the
minimum-mass constraint in low dimension and by the growth of the BIC penalty
with $d$ in high dimension, not by the cluster's shape.

\paragraph{Implications.}
Two consequences follow for ASM as a whole. First, in low dimension and at the
sample sizes of a pooled dataset, a single true cluster reaches the merge stage
as tens to hundreds of components, so the accuracy of the final count rests on
the merge rule and on the choice of $\alpha$; from Appendix~\ref{app:bic}, the
fragments of a cluster with anisotropy $r$ are reunited only when
$\alpha \gtrsim 0.8\sqrt{r}$. Second, the number of components grows with the
number of points per cluster, so smaller per-client datasets produce fewer
spurious components. This is the sense in which ASM is better suited to the
federated regime of moderately sized clients than to the centralized limit.

\section{Analysis of ASM Merge}

\begin{table}[h]
\centering
\small
\caption{Merge ablation for ASM-S on pooled data over a range of merge thresholds $\alpha$. Held-out ARI
(mean over 10 seeds, 42--51) with the mean number of components $\hat K$ below it; ``no merge'' is the split tree
itself. Bold: the $\alpha$ selected by silhouette. $^{\uparrow}$/$^{\downarrow}$: ARI significantly higher/lower than
without the merge (Wilcoxon signed-rank, paired by seed, $p<0.05$). At $\alpha=0.25$ the merge never fires on any dataset
(identical to no merge). $^{\ast}$MNIST's selected $\alpha=1.1$ is not a column: ARI 0.162, $\hat K$ 81.0.}
\label{tab:fv_merge_alpha_sweep_pooled}
\resizebox{\ifdim\width>\linewidth\linewidth\else\width\fi}{!}{%
\begin{tabular}{lcccccccc}
\toprule
 & & \multicolumn{7}{c}{Merge threshold $\alpha$} \\
\cmidrule(lr){3-9}
Dataset & no merge & 0.5 & 0.75 & 1.0 & 1.25 & 1.5 & 2.0 & 3.0 \\
\midrule
Waveform & \makecell{0.083 \\ {}$\hat K$ 28.8} & \makecell{0.083 \\ {}$\hat K$ 28.8} & \makecell{0.083 \\ {}$\hat K$ 28.8} & \makecell{0.083 \\ {}$\hat K$ 28.8} & \makecell{0.208$^{\uparrow}$ \\ {}$\hat K$ 11.1} & \makecell{0.303$^{\uparrow}$ \\ {}$\hat K$ 6.2} & \makecell{\textbf{0.256}$^{\uparrow}$ \\ {}$\hat K$ 3.0} & \makecell{0.050 \\ {}$\hat K$ 1.4} \\
\midrule
Abalone & \makecell{0.076 \\ {}$\hat K$ 16.3} & \makecell{0.076 \\ {}$\hat K$ 16.3} & \makecell{0.076 \\ {}$\hat K$ 16.3} & \makecell{0.076 \\ {}$\hat K$ 16.2} & \makecell{0.077 \\ {}$\hat K$ 15.4} & \makecell{0.077 \\ {}$\hat K$ 9.9} & \makecell{\textbf{0.087}$^{\uparrow}$ \\ {}$\hat K$ 7.5} & \makecell{0.086$^{\uparrow}$ \\ {}$\hat K$ 3.5} \\
\midrule
FrogA & \makecell{0.225 \\ {}$\hat K$ 15.5} & \makecell{0.225 \\ {}$\hat K$ 15.5} & \makecell{0.225 \\ {}$\hat K$ 15.5} & \makecell{0.222 \\ {}$\hat K$ 15.2} & \makecell{0.219$^{\downarrow}$ \\ {}$\hat K$ 14.6} & \makecell{0.215$^{\downarrow}$ \\ {}$\hat K$ 14.0} & \makecell{\textbf{0.273}$^{\uparrow}$ \\ {}$\hat K$ 10.6} & \makecell{0.497$^{\uparrow}$ \\ {}$\hat K$ 6.7} \\
\midrule
FrogB & \makecell{0.139 \\ {}$\hat K$ 15.6} & \makecell{0.139 \\ {}$\hat K$ 15.6} & \makecell{0.139 \\ {}$\hat K$ 15.6} & \makecell{0.136 \\ {}$\hat K$ 14.9} & \makecell{0.134 \\ {}$\hat K$ 14.7} & \makecell{0.127$^{\downarrow}$ \\ {}$\hat K$ 13.9} & \makecell{\textbf{0.153} \\ {}$\hat K$ 10.8} & \makecell{0.275$^{\uparrow}$ \\ {}$\hat K$ 6.2} \\
\midrule
MNIST$^{\ast}$ & \makecell{0.090 \\ {}$\hat K$ 150.9} & \makecell{0.090 \\ {}$\hat K$ 150.9} & \makecell{0.090 \\ {}$\hat K$ 150.9} & \makecell{0.112$^{\uparrow}$ \\ {}$\hat K$ 121.3} & \makecell{0.200$^{\uparrow}$ \\ {}$\hat K$ 27.2} & \makecell{0.136$^{\uparrow}$ \\ {}$\hat K$ 16.0} & \makecell{0.002$^{\downarrow}$ \\ {}$\hat K$ 1.2} & \makecell{0.000$^{\downarrow}$ \\ {}$\hat K$ 1.0} \\
\midrule
FMNIST & \makecell{0.070 \\ {}$\hat K$ 152.5} & \makecell{0.070 \\ {}$\hat K$ 152.5} & \makecell{0.070 \\ {}$\hat K$ 152.5} & \makecell{0.070 \\ {}$\hat K$ 152.5} & \makecell{0.070 \\ {}$\hat K$ 152.5} & \makecell{0.070 \\ {}$\hat K$ 152.5} & \makecell{0.070 \\ {}$\hat K$ 152.5} & \makecell{0.071$^{\uparrow}$ \\ {}$\hat K$ 149.9} \\
\midrule
EMNIST & \makecell{0.180 \\ {}$\hat K$ 193.2} & \makecell{0.180 \\ {}$\hat K$ 193.2} & \makecell{0.180 \\ {}$\hat K$ 193.2} & \makecell{0.180 \\ {}$\hat K$ 193.1} & \makecell{0.189$^{\uparrow}$ \\ {}$\hat K$ 176.7} & \makecell{0.230$^{\uparrow}$ \\ {}$\hat K$ 76.6} & \makecell{0.057$^{\downarrow}$ \\ {}$\hat K$ 21.6} & \makecell{0.000$^{\downarrow}$ \\ {}$\hat K$ 1.0} \\
\midrule
CIFAR-10 & \makecell{0.120 \\ {}$\hat K$ 112.5} & \makecell{0.120 \\ {}$\hat K$ 112.5} & \makecell{0.120 \\ {}$\hat K$ 112.5} & \makecell{0.120 \\ {}$\hat K$ 112.5} & \makecell{0.121 \\ {}$\hat K$ 112.1} & \makecell{0.122$^{\uparrow}$ \\ {}$\hat K$ 110.8} & \makecell{\textbf{0.156}$^{\uparrow}$ \\ {}$\hat K$ 87.7} & \makecell{0.164$^{\uparrow}$ \\ {}$\hat K$ 30.9} \\
\bottomrule
\end{tabular}}
\end{table}

\section{Our \fedgem\ implementation and the released code}
\label{app:repro}
\paragraph{Why a separate implementation.}
\asm\ is evaluated inside \fedgem, so every result depends on the aggregator being implemented as specified. Running
the released code unmodified, we found three steps that we implement differently: two in which the code departs from
the algorithm of~\citet{ibrahim2026fedgem}, and one in which paper and code agree but, as far as we can tell, contain
a sign error.
(i) The per-round uncertainty radius is meant to be found by bisection. In the released code the condition of the
bisection loop is never satisfied on entry, so the radius always keeps its initial value $\|\hat M-\theta\|^2/2$.
Our implementation runs the bisection. Run unmodified on MNIST, the released code reproduces the published global
cardinality almost exactly ($\hat K = 13.6$ against the published $13.6$), which indicates that the published numbers
were obtained with this unrefined radius.
(ii) When the uncertainty balls $B(\theta_1,r_1)$ and $B(\theta_2,r_2)$ of two clients overlap, the server moves both
components to a point $\nu$ intended to lie in their intersection. Algorithm~2 of the paper and the released code
step from $\theta_1$ in the direction $\theta_1-\theta_2$, away from the second ball; we step in the direction
$\theta_2-\theta_1$, toward it. Over 20{,}000 random overlapping pairs in 2 to 64 dimensions, our $\nu$ lies in both
balls in every case and the published formula in $5.9\%$ of cases, only when the second ball already contains
$\theta_1$ with room to spare.
(iii) The released code stops early when the means change by less than $10^{-3}$; we run a fixed $T=10$ rounds
(the released FMNIST script uses $T=20$).
As in the released code, components of the same client are never grouped together.

\paragraph{Metric and results.}
The released code reports the mean over clients of each client's ARI under its own components (personalised ARI).
Our primary metric is the ARI of the global model on the pooled test set, and we report both.
Table~\ref{tab:repro} compares, on the personalised metric, the published results, the released code as run by
us
(8--10 runs on seven datasets) and our implementation (\oraclek, ten repetitions on all eight). The implementations
differ in accuracy in both directions: ours is higher on Waveform, Abalone, FMNIST and CIFAR-10 and lower on
FrogA
and FrogB. Ours, however, estimates the global cardinality much more closely. Its $|\hat K-K^\star|$ is smaller than
that of the published results on seven of the eight datasets, for example FrogA $7.9$ against $23.9$ with
$K^\star=10$ and EMNIST $49.5$ against $58.7$ with $K^\star=47$. The exception is CIFAR-10, where both overshoot
($40.8$ against $37.7$). It is also smaller than that of the released code on six of the seven datasets on which we
ran it; the exception is Waveform. On FMNIST the released script, run with its own $T=20$, reaches $\hat K = 90$;
with $T=10$ as in ours it gives $9.1$. Because \oraclek\ run through the same
implementation, the comparisons of Section~\ref{sec:experiments2} do not depend on these differences.

\begin{table}[t]
\centering
\small
\caption{\fedgem\ with the true local cardinalities: personalised ARI $\pm$ 95\% CI (t-distribution) and, below it,
the mean global $\hat K$. \emph{Published}: Table~1 of \citet{ibrahim2026fedgem}, CI from the reported SD with
$n = 50$. \emph{Released code}: the authors' unmodified \fedgem\ code run by us (8--10 runs; FMNIST with the script's
$T = 20$; not run on CIFAR-10). \emph{Ours}: our implementation, ten repetitions (seeds 42--51); last column its
global ARI. Protocols also differ in partition, number of repetitions and feature standardisation.}
\label{tab:repro}
\resizebox{\ifdim\width>\linewidth\linewidth\else\width\fi}{!}{%
\begin{tabular}{lccccc}
\toprule
 &  & \multicolumn{3}{c}{personalised ARI ($\hat K$)} & \\
\cmidrule(lr){3-5}
Dataset & $K^\ast$ & published & released code & ours & ours, global ARI \\
\midrule
Waveform & 3 & \makecell{0.335 $\pm$ 0.022 \\ {}$\hat K$ 4.4} & \makecell{0.318 $\pm$ 0.078 \\ {}$\hat K$ 3.7} & \makecell{0.405 $\pm$ 0.063 \\ {}$\hat K$ 2.0} & 0.339 $\pm$ 0.024 \\
Abalone & 8 & \makecell{0.138 $\pm$ 0.016 \\ {}$\hat K$ 12.1} & \makecell{0.131 $\pm$ 0.032 \\ {}$\hat K$ 15.2} & \makecell{0.174 $\pm$ 0.036 \\ {}$\hat K$ 6.0} & 0.103 $\pm$ 0.007 \\
FrogA & 10 & \makecell{0.552 $\pm$ 0.037 \\ {}$\hat K$ 23.9} & \makecell{0.626 $\pm$ 0.063 \\ {}$\hat K$ 31.4} & \makecell{0.489 $\pm$ 0.055 \\ {}$\hat K$ 7.9} & 0.601 $\pm$ 0.104 \\
FrogB & 8 & \makecell{0.468 $\pm$ 0.033 \\ {}$\hat K$ 20.0} & \makecell{0.481 $\pm$ 0.128 \\ {}$\hat K$ 22.9} & \makecell{0.318 $\pm$ 0.089 \\ {}$\hat K$ 5.9} & 0.451 $\pm$ 0.093 \\
MNIST & 10 & \makecell{0.452 $\pm$ 0.014 \\ {}$\hat K$ 13.6} & \makecell{0.481 $\pm$ 0.033 \\ {}$\hat K$ 13.6} & \makecell{0.490 $\pm$ 0.014 \\ {}$\hat K$ 9.4} & 0.446 $\pm$ 0.048 \\
FMNIST & 10 & \makecell{0.287 $\pm$ 0.016 \\ {}$\hat K$ 17.6} & \makecell{0.195 $\pm$ 0.031 \\ {}$\hat K$ 90.0} & \makecell{0.362 $\pm$ 0.009 \\ {}$\hat K$ 8.0} & 0.252 $\pm$ 0.027 \\
EMNIST & 47 & \makecell{0.285 $\pm$ 0.006 \\ {}$\hat K$ 58.7} & \makecell{0.322 $\pm$ 0.026 \\ {}$\hat K$ 57.7} & \makecell{0.332 $\pm$ 0.004 \\ {}$\hat K$ 49.5} & 0.257 $\pm$ 0.011 \\
CIFAR-10 & 10 & \makecell{0.286 $\pm$ 0.009 \\ {}$\hat K$ 37.7} & -- & \makecell{0.546 $\pm$ 0.019 \\ {}$\hat K$ 40.8} & 0.442 $\pm$ 0.017 \\
\bottomrule
\end{tabular}}
\end{table}


\section{Implementation and protocol details}
\label{app:details} 

\begin{table}[h]
\centering
\caption{Datasets and federation settings. $\overline{K_g}$: mean number of classes per client.}
\label{tab:datasets}
\small
\begin{adjustbox}{max width=\linewidth}
\begin{tabular}{lrrrrrrl}
\toprule
Dataset & $N$ & $d$ & $K^\star$ & $G$ & train/test & $\overline{K_g}$ & features \\
\midrule
Waveform & 5,000 & 21 & 3 & 5 & 70/30 & 2.0 & raw, standardised \\
Abalone & 4,177 & 7 & 8 & 5 & 70/30 & 4.5 & raw (sex removed), standardised \\
FrogA & 7,195 & 21 & 10 & 5 & 70/30 & 6.3 & raw, standardised \\
FrogB & 7,195 & 21 & 8 & 5 & 70/30 & 4.3 & raw, standardised \\
MNIST & 70,000 & 10 & 10 & 100 & 70/30 & 6.1 & VAE embedding \\
FMNIST & 70,000 & 64 & 10 & 100 & 70/30 & 6.1 & VAE embedding \\
EMNIST & 131,600 & 16 & 47 & 25 & 50/50 & 26.1 & VAE embedding \\
\bottomrule
\end{tabular}
\end{adjustbox}
\end{table}

\paragraph{Data.}
The tabular features are standardised; as in the released loader of~\citet{ibrahim2026fedgem}, the categorical
sex
attribute of Abalone is removed, and its age bins yield $K^\star=8$. The image datasets are used as embeddings
without
standardisation:MNIST ($d=10$) uses the released VAE embeddings;
FMNIST ($d=64$) and EMNIST ($d=16$) use VAE embeddings that we trained with the architecture specified by~\citet{ibrahim2026fedgem};
CIFAR-10 ($d=64$) uses the released Barlow Twins embeddings passed through a second-stage VAE that we retrained, because the original was not released.
FrogA and FrogB share one feature matrix and differ only in their labels. Every evaluation seed (42--51) draws a
stratified train/test split: 70/30, and 50/50 for EMNIST. Every class is held by at least two clients.

\paragraph{Local EM and comparison methods.}
\begin{itemize}
  \item \emph{ASM-I} (identity covariance) uses a dedicated EM implementation with fixed $\Sigma=\mI$ and
estimated
  weights.
  \item \emph{ASM-S} (spherical covariance) uses scikit-learn's \texttt{GaussianMixture}~\citep{pedregosa2011},
with
  the perturbed means as initial means.
  \item \emph{Merge rule.} All \asm\ runs, under both ASM-I and ASM-S, use the linear merge rule: two components
are
  merged when $\|\mu_i-\mu_j\|\le\alpha(\sigma_i+\sigma_j)$, with $\sigma=\sqrt{\operatorname{tr}\Sigma/d}$.
  \item \emph{DP-GMM:} scikit-learn \texttt{BayesianGaussianMixture} with spherical covariance, truncation level
200 and
  at most 200 iterations. Components below the weight threshold are removed, and their points are reassigned to
the
  nearest remaining mean.
  \item \emph{X-Means and G-Means:} the \texttt{pyclustering} implementations with at most 200 clusters.
\end{itemize}
Every Phase~1 method therefore has the same capacity of 200 components.

\paragraph{Hyperparameter grids.}
\begin{center}
\begin{adjustbox}{max width=\linewidth}
\begin{tabular}{ll}
\toprule
Setting & Grid \\
\midrule
\asm\ (ASM-I / ASM-S), real data (federated: selected per client; pooled: one configuration; 120 configurations)
& $A\in\{20,50,80,120,200\}$, $\delta\in\{0.3,0.5,0.7,1.0\}$, $\alpha\in\{0.25,0.5,0.75,1.0,1.5,2.0\}$\\
\asm\ (ASM-I / ASM-S), MNIST (140 configurations) & as above, with
$\alpha\in\{0.25,0.5,0.75,0.9,1.0,1.1,1.2\}$\\
DP-GMM (truncation level 200; 24 configurations) & concentration $\in\{0.001,0.01,0.1,1,10,100\}$, weight
threshold $\in\{0.001,0.01,0.03,0.1\}$\\
$\upsilon_g$, tabular data & Waveform $\{1,5,10\}$; Abalone $\{5000,7000,9000\}$; FrogA, FrogB
$\{10^4,10^5,10^6\}$\\
$\upsilon_g$, image embeddings & fixed: MNIST 2, FMNIST 300, EMNIST 2, CIFAR-10 10\\
\bottomrule
\end{tabular}
\end{adjustbox}
\end{center}
With per-client selection, each client chooses its own configuration from the grid by the silhouette of its
Phase~1
clustering on its own held-out points.

\emph{Time limits.} Each \fedgem\ run used to select $\upsilon_g$ was limited to 150\,s, and values whose runs
timed
out on both tuning seeds were excluded from selection. Evaluation runs were limited to 3{,}600\,s. G-Means
exceeded
this limit on 5 of 10 seeds for CIFAR-10 on Regime A's partition; these were filled from an identical rerun.
It also exceeded the limit on 1 seed for EMNIST on Regime B's partition, which is reported over the remaining 9. No other evaluation run timed out.

\emph{CIFAR-10 $\upsilon_g$.} The value of 10 differs from the paper's 20. The paper did not release itssecond-stage
VAE, so we retrained it, and the retrained features have a different scale (a standard deviation of about 2.2 per
dimension, against about 1 for the other embeddings). On these features the silhouette differences across the
$\upsilon_g$ grid were within noise. We therefore fixed $\upsilon_g$ at the value at which our implementation's
global cardinality is closest to the published one.

\paragraph{Selection score.}
Each client's training data are split at random into fitting (80\%) and validation (20\%) sets. Every client fits Phase~1 with each configuration on its fitting set and keeps the configuration with the largest silhouette of its own validation points under its own components (ties:smaller $\hat K_g$, then grid order). No information is pooled across clients.

\paragraph{Pooled regime.}
With a single client holding all data, configurations are selected on a random subsample of at most 30{,}000 points
split 80/20 (tuning seeds 1000, 1001). The score is the silhouette of the validation points, each assigned to the
nearest component mean. Every method is then fitted on the training part of each evaluation split and scored by the ARI of the held-out test points assigned to the nearest component mean.

\section{Regime A and B}
\label{app:partitions}
\begin{figure}[!htbp]
\centering
\includegraphics[width=\linewidth]{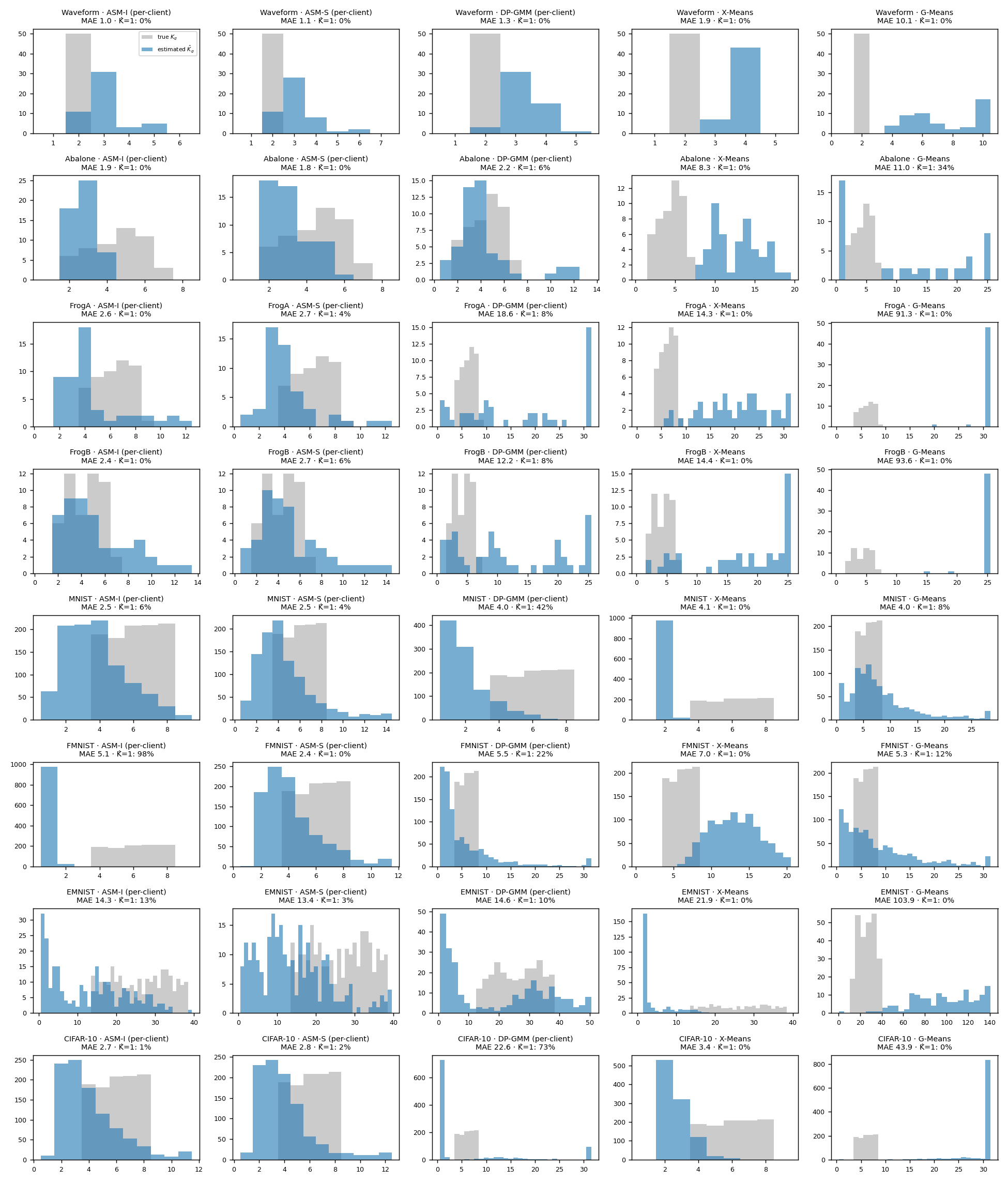}
\caption{Local cardinalities under Regime A: true $K_g$ (grey) and estimated $\hat K_g$ (blue) for
every client over ten repetitions. Rows are datasets; columns are, from left to right, \asmid, \asmsph, DP-GMM,
X-Means and G-Means. Panel titles give the mean absolute error of $\hat K_g$ and the share of clients with
$\hat K_g = 1$.}
\label{fig:localk-hist-all}
\end{figure}

\begin{figure}[!htbp]
\centering
\includegraphics[width=\linewidth]{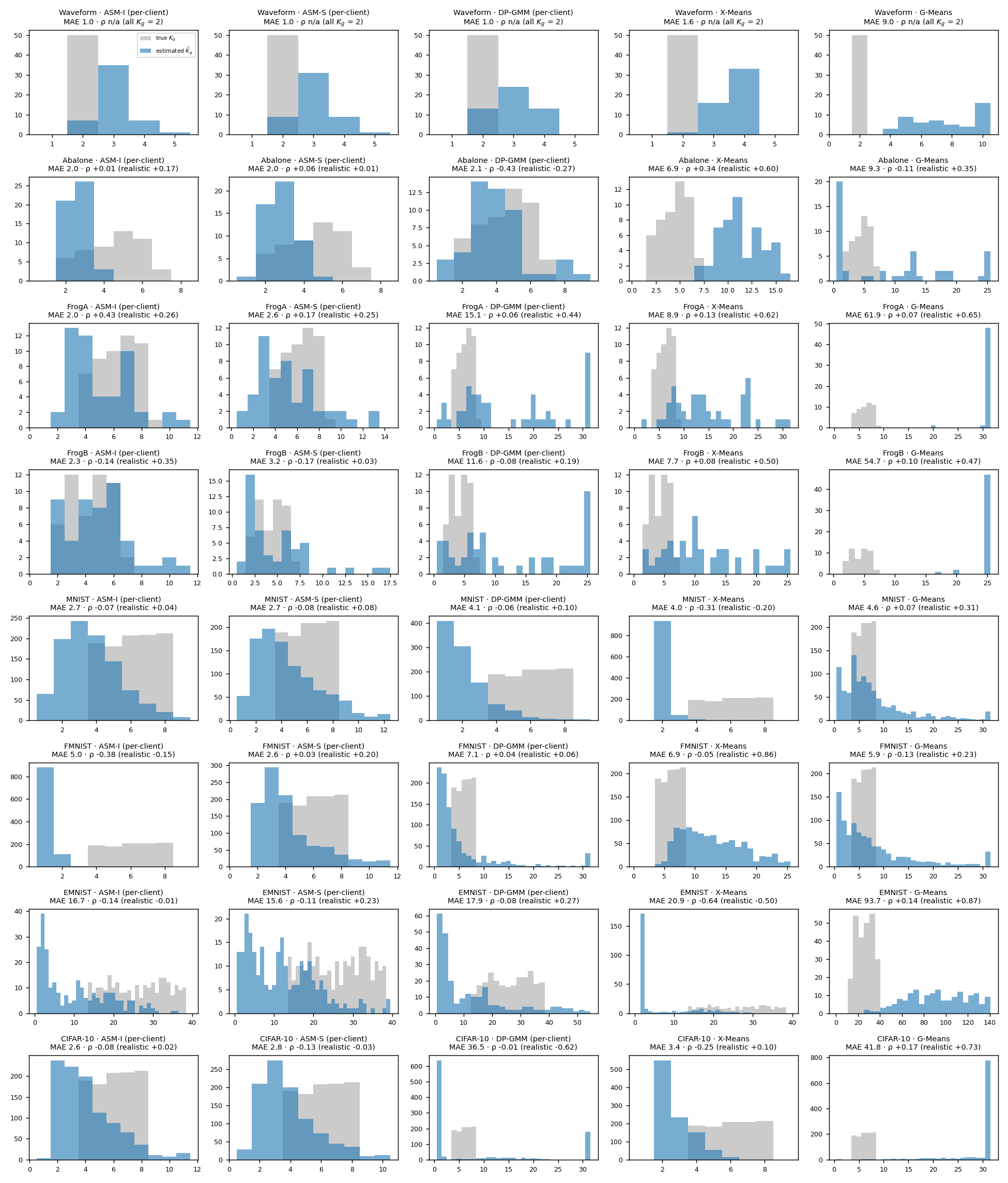}
\caption{Local cardinalities under the Regime B, laid out as in Figure~\ref{fig:localk-hist-all}. Panel
titles give the mean absolute error of $\hat K_g$ and the Spearman correlation $\rho$ between $\hat K_g$ and $K_g$,
with the value under the Regime A in brackets.}
\label{fig:localk-hist}
\end{figure}

Figures~\ref{fig:localk-hist-all} and~\ref{fig:localk-hist} show what Phase~1 hands to Phase~2 on every client. Each
row is one dataset and each column one method. The grey histogram gives the true local cardinalities $K_g$ of all clients over the ten repetitions, and the blue histogram gives the estimates $\hat K_g$ of the same clients, the method is accurate when the two histograms overlap. 

\paragraph{Train/test split and clients.}
For every evaluation seed, each dataset is first split into a training and a held-out test set, stratified by class. Only the training set is distributed to clients. The number of clients $G$ is fixed per
dataset: $G=5$ for Waveform, Abalone, FrogA and FrogB, $G=25$ for EMNIST and $G=100$ for MNIST, FMNIST and  CIFAR-10. A
client's true local cardinality $K_g$ is the number of distinct clusters present among the data points it receives.       
\paragraph{Regime A.}                                     
\begin{enumerate}                                                    \item \emph{Class sets.} Each client draws a number of classes uniformly from $\{\lfloor 0.55K\rceil - j, \dots, \lfloor 0.55K\rceil + j\}$ with $j=\max(1,\lfloor 0.25K\rceil)$, clipped to
$[2, K-1]$, and then that many classes uniformly at random without replacement. Every client therefore holds a strict subset of the classes (for Waveform, with $K=3$, every client holds exactly two).
\item \emph{Coverage.} Any class held by fewer than two clients is added to randomly chosen clients that do not yet
hold it and hold fewer than $K-1$ classes, until every class has at least two owners.
\item \emph{Allocation.} The training points of each class are shuffled and divided into equal parts among the clients that hold the class. Clients are disjoint and every training point is assigned.
\end{enumerate}
A draw is accepted if every client holds between $2$ and $K-1$ classes and at least $100$ points; otherwise it is
redrawn (at most 200 times, keeping the draw with the largest minimum client size). Because a client receives an equal
share of each of its classes, its size grows with its number of classes: the Spearman correlation between $N_g$and
$K_g$ is $0.98$--$0.99$ on the image datasets and $0.85$ on Abalone, and the number of points per class, $N_g/K_g$, varies little across clients (coefficient of variation $0.04$ on the image datasets). The correlation is weaker on FrogA ($0.52$) and FrogB ($0.36$), whose classes differ strongly in size, and undefined on Waveform, where every $K_g = 2$. This partition is representative of practice, but a method that counts points could appear to count clusters.

\paragraph{Regime B.}
Regime B's partition is derived from the Regime A's partition of the same seed and removes the link between size and cardinality while keeping each client's class set, and hence its true $K_g$.
\begin{enumerate}
  \item \emph{Target sizes.} Let $N_{\mathrm{med}}$ be the median client size of Regime A. Each client
  draws a target size $T_g = N_{\mathrm{med}}\, e^{u_g}$ with $u_g \sim \mathcal{U}[\log 0.5, \log 2]$, independently of
  $K_g$, i.e.\ log-uniformly over the fourfold range $[0.5N_{\mathrm{med}}, 2N_{\mathrm{med}}]$.
  \item \emph{Allocation.} Each client requests $T_g/K_g$ points from each of its classes. When the requests for a class
  exceed its training points, all requests for that class are scaled down by the same factor. Each client then
  receives $\max(2, \mathrm{round}(\cdot))$ points of each of its classes, drawn without replacement from the shuffled
  class pool, so clients remain disjoint; unlike Regime A, not every training point need be used.
  \item \emph{Fallback.} A client that would receive fewer than 60 points keeps its Regime A data.
\end{enumerate}
Regime B's partition removes the size shortcut: on the image datasets the correlation between $N_g$ and $K_g$ is
statistically equivalent to zero ($-0.01$ to $0.04$), the coefficient of variation of $N_g/K_g$ rises from $0.04$ to
$0.48$--$0.50$, and the elasticity $b$ in $\log N_g = a + b\log K_g$ falls from $1.00$--$1.03$ to $-0.03$--$0.05$. A side effect is that clients with more classes now hold fewer points per class.

\paragraph{Method settings on the two partitions.}
On both partitions each client selects its own Phase~1 configuration on its own data (Appendix~\ref{app:details}); only
the \fedgem\ setting $\upsilon_g$ is carried over from Regime A's tuning. \oraclek\ is given the true class count of
each client. All other protocol details (evaluation seeds 42--51, held-out test set, statistics) are as in
Appendix~\ref{app:details}.

\begin{table}[h]
\centering
\small
\caption{Full federated pipeline, Regime B, per-client hyperparameter selection. Global ARI $\pm$ 95\% CI half-width (t, n = 10 seeds 42--51) on the Regime B; second line $\hat K$, $|\Delta K|$ = mean |$\hat K$ $-$ $K^\ast$|. Mean row = unweighted mean over 8 datasets.}
\label{tab:fv_federated_decoupled}
\resizebox{\textwidth}{!}{%
\begin{tabular}{lcccccc}
\toprule
Dataset & ASM-S & DP-GMM & X-Means & G-Means & Oracle-K \\
\midrule
Waveform & \makecell{0.280 $\pm$ 0.022 \\ $\hat K$ 4.4 $\cdot$ $|\Delta K|$ 1.4} & \makecell{0.252 $\pm$ 0.027 \\ $\hat K$ 4.1 $\cdot$ $|\Delta K|$ 1.1} & \makecell{0.271 $\pm$ 0.032 \\ $\hat K$ 5.2 $\cdot$ $|\Delta K|$ 2.2} & \makecell{0.138 $\pm$ 0.035 \\ $\hat K$ 21.3 $\cdot$ $|\Delta K|$ 18.3} & \makecell{0.340 $\pm$ 0.011 \\ $\hat K$ 2.0 $\cdot$ $|\Delta K|$ 1.0} \\
Abalone & \makecell{0.092 $\pm$ 0.010 \\ $\hat K$ 3.8 $\cdot$ $|\Delta K|$ 4.2} & \makecell{0.097 $\pm$ 0.014 \\ $\hat K$ 6.5 $\cdot$ $|\Delta K|$ 1.7} & \makecell{0.100 $\pm$ 0.008 \\ $\hat K$ 14.4 $\cdot$ $|\Delta K|$ 6.4} & \makecell{0.071 $\pm$ 0.012 \\ $\hat K$ 28.4 $\cdot$ $|\Delta K|$ 20.4} & \makecell{0.101 $\pm$ 0.007 \\ $\hat K$ 6.0 $\cdot$ $|\Delta K|$ 2.0} \\
FrogA & \makecell{0.651 $\pm$ 0.123 \\ $\hat K$ 9.1 $\cdot$ $|\Delta K|$ 2.3} & \makecell{0.384 $\pm$ 0.169 \\ $\hat K$ 53.2 $\cdot$ $|\Delta K|$ 43.2} & \makecell{0.494 $\pm$ 0.169 \\ $\hat K$ 24.9 $\cdot$ $|\Delta K|$ 14.9} & \makecell{0.304 $\pm$ 0.116 \\ $\hat K$ 99.8 $\cdot$ $|\Delta K|$ 89.8} & \makecell{0.588 $\pm$ 0.138 \\ $\hat K$ 7.9 $\cdot$ $|\Delta K|$ 2.1} \\
FrogB & \makecell{0.442 $\pm$ 0.083 \\ $\hat K$ 10.4 $\cdot$ $|\Delta K|$ 3.0} & \makecell{0.255 $\pm$ 0.055 \\ $\hat K$ 33.2 $\cdot$ $|\Delta K|$ 25.2} & \makecell{0.329 $\pm$ 0.094 \\ $\hat K$ 20.9 $\cdot$ $|\Delta K|$ 12.9} & \makecell{0.190 $\pm$ 0.041 \\ $\hat K$ 89.5 $\cdot$ $|\Delta K|$ 81.5} & \makecell{0.385 $\pm$ 0.096 \\ $\hat K$ 5.9 $\cdot$ $|\Delta K|$ 2.1} \\
MNIST & \makecell{0.420 $\pm$ 0.032 \\ $\hat K$ 14.7 $\cdot$ $|\Delta K|$ 4.7} & \makecell{0.408 $\pm$ 0.025 \\ $\hat K$ 11.1 $\cdot$ $|\Delta K|$ 1.9} & \makecell{0.248 $\pm$ 0.051 \\ $\hat K$ 4.6 $\cdot$ $|\Delta K|$ 5.4} & \makecell{0.273 $\pm$ 0.048 \\ $\hat K$ 53.6 $\cdot$ $|\Delta K|$ 43.6} & \makecell{0.418 $\pm$ 0.057 \\ $\hat K$ 8.7 $\cdot$ $|\Delta K|$ 1.3} \\
FMNIST & \makecell{0.274 $\pm$ 0.022 \\ $\hat K$ 14.2 $\cdot$ $|\Delta K|$ 4.4} & \makecell{0.172 $\pm$ 0.027 \\ $\hat K$ 114.3 $\cdot$ $|\Delta K|$ 104.3} & \makecell{0.283 $\pm$ 0.021 \\ $\hat K$ 25.6 $\cdot$ $|\Delta K|$ 15.6} & \makecell{0.247 $\pm$ 0.024 \\ $\hat K$ 46.6 $\cdot$ $|\Delta K|$ 36.6} & \makecell{0.250 $\pm$ 0.025 \\ $\hat K$ 8.0 $\cdot$ $|\Delta K|$ 2.0} \\
EMNIST & \makecell{0.240 $\pm$ 0.011 \\ $\hat K$ 44.7 $\cdot$ $|\Delta K|$ 7.1} & \makecell{0.258 $\pm$ 0.011 \\ $\hat K$ 71.2 $\cdot$ $|\Delta K|$ 24.2} & \makecell{0.264 $\pm$ 0.021 \\ $\hat K$ 47.0 $\cdot$ $|\Delta K|$ 7.8} & \makecell{0.149 $\pm$ 0.008 \\ $\hat K$ 364.8 $\cdot$ $|\Delta K|$ 317.8 $\cdot$ n = 9} & \makecell{0.242 $\pm$ 0.007 \\ $\hat K$ 46.4 $\cdot$ $|\Delta K|$ 1.2} \\
CIFAR-10 & \makecell{0.379 $\pm$ 0.037 \\ $\hat K$ 31.7 $\cdot$ $|\Delta K|$ 21.7} & \makecell{0.096 $\pm$ 0.008 \\ $\hat K$ 199.4 $\cdot$ $|\Delta K|$ 189.4} & \makecell{0.403 $\pm$ 0.040 \\ $\hat K$ 21.8 $\cdot$ $|\Delta K|$ 11.8} & \makecell{0.155 $\pm$ 0.013 \\ $\hat K$ 111.3 $\cdot$ $|\Delta K|$ 101.3} & \makecell{0.415 $\pm$ 0.040 \\ $\hat K$ 43.6 $\cdot$ $|\Delta K|$ 33.6} \\
\bottomrule
\end{tabular}}
\end{table}


\subsection{Points per cluster in Regime A and B}
\label{app:points-per-cluster}

\begin{figure}[h]
\centering
\includegraphics[width=0.8\linewidth]{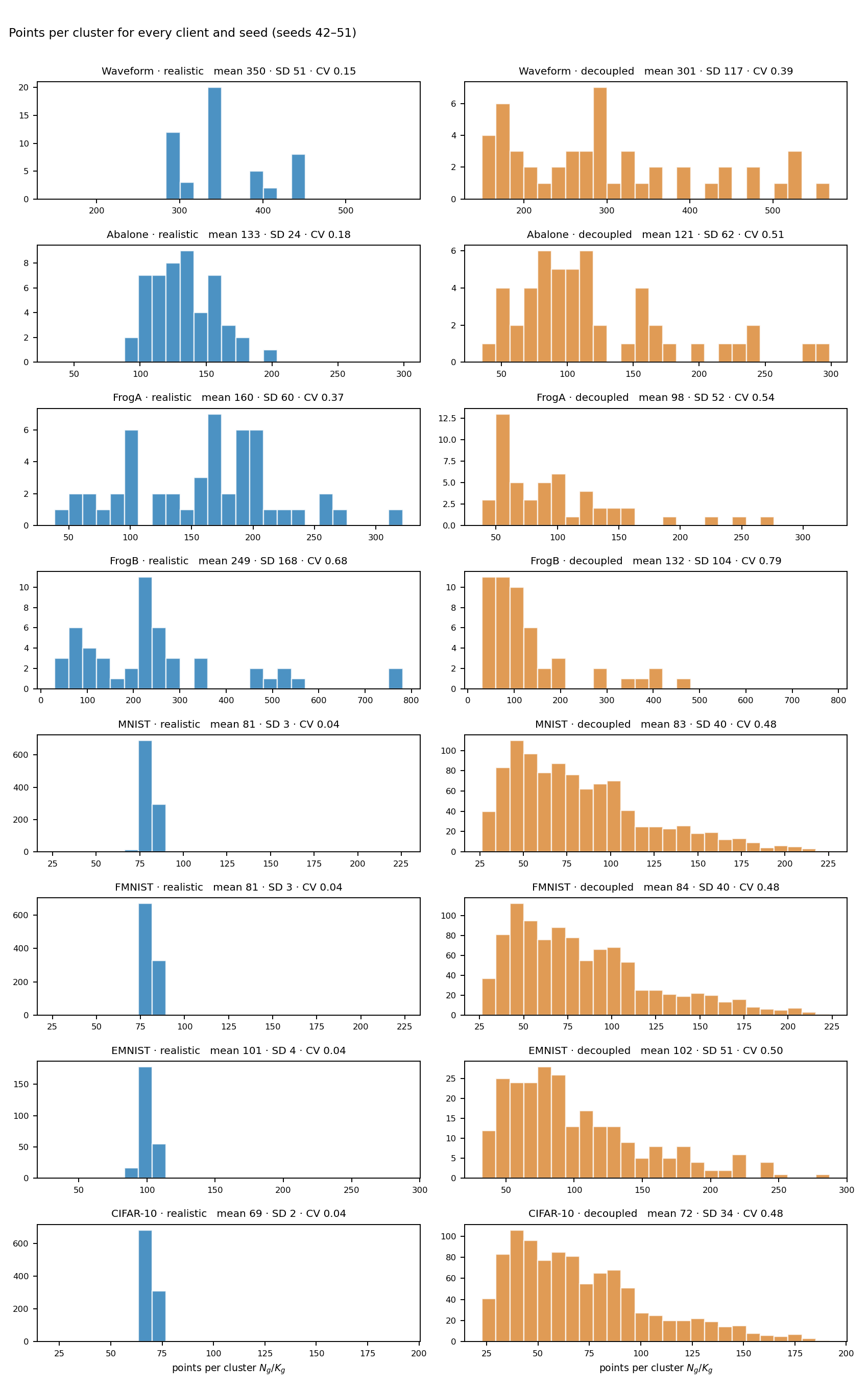}
\caption{Points per cluster $N_g/K_g$ for every client and seed (seeds 42--51), Regime A (left) and Regime B (right), with the mean, SD and CV of each distribution.}
\label{fig:points-per-cluster}
\end{figure}

Regime B's partition is meant to remove the link between a client's size and its number of classes. We check this
directly with a simple quantity, the number of points a client holds per class,
\[
  \text{points per cluster} \;=\; N_g / K_g .
\]
If client size is proportional to cardinality, as the construction of Regime A's partition implies, $N_g/K_g$ is
nearly the same for every client; if size is unrelated to cardinality, it varies widely. We compute it for every client
of every evaluation seed (42--51) on both partitions, using the partitions of the main experiments, and summarise its
spread by the standard deviation (SD) and the coefficient of variation, $\mathrm{CV} = \mathrm{SD}/\text{mean}$, which
allows datasets with different class sizes to be compared.

\paragraph{Descriptive results.}
Table~\ref{tab:fv_points_per_cluster} and Figure~\ref{fig:points-per-cluster} report the distribution over all
clients and seeds. On the four image datasets Regime A's partition gives almost every client the same number of
points per class: the CV is $0.04$ (SD $2$--$4$ points around means of $69$--$101$), and the Spearman correlation between
$N_g$ and $K_g$ is $0.98$--$0.99$. In Regime B's the mean is essentially unchanged, but the spread grows
about twelve fold (CV $0.48$--$0.50$, SD $34$--$51$), and the correlation between $N_g$ and $K_g$ vanishes ($-0.01$ to
$0.04$). The tabular datasets behave in the same direction but start from a larger spread in Regime A
(CV $0.15$--$0.68$), for the reason discussed below.

\begin{table}[t]
\centering
\small
\caption{Points per cluster $N_g/K_g$ over every client and seed (seeds 42--51) of the realistic and decoupled
partitions: mean, standard deviation, coefficient of variation (SD/mean) and Spearman correlation between client
size $N_g$ and cardinality $K_g$ (``--'': every Waveform client has $K_g = 2$). A near-constant $N_g/K_g$ (low CV)
means client size is proportional to cardinality.}
\label{tab:fv_points_per_cluster}
\resizebox{\ifdim\width>\linewidth\linewidth\else\width\fi}{!}{%
\begin{tabular}{lc cccc cccc}
\toprule
 & & \multicolumn{4}{c}{realistic} & \multicolumn{4}{c}{decoupled} \\
\cmidrule(lr){3-6} \cmidrule(lr){7-10}
Dataset & cases & mean & SD & CV & $\rho(N_g, K_g)$ & mean & SD & CV & $\rho(N_g, K_g)$ \\
\midrule
Waveform & 50 & 350 & 51 & 0.15 & -- & 301 & 117 & 0.39 & -- \\
Abalone & 50 & 133 & 24 & 0.18 & +0.85 & 121 & 62 & 0.51 & +0.14 \\
FrogA & 50 & 160 & 60 & 0.37 & +0.52 & 98 & 52 & 0.54 & +0.08 \\
FrogB & 50 & 249 & 168 & 0.68 & +0.36 & 132 & 104 & 0.79 & -0.08 \\
MNIST & 1000 & 81 & 3 & 0.04 & +0.98 & 83 & 40 & 0.48 & -0.01 \\
FMNIST & 1000 & 81 & 3 & 0.04 & +0.98 & 84 & 40 & 0.48 & -0.01 \\
EMNIST & 250 & 101 & 4 & 0.04 & +0.99 & 102 & 51 & 0.50 & +0.04 \\
CIFAR-10 & 1000 & 69 & 2 & 0.04 & +0.98 & 72 & 34 & 0.48 & -0.01 \\
\bottomrule
\end{tabular}}
\end{table}

\paragraph{Statistical tests.}
Clients of the same seed share one partition and are therefore not independent, so every test below uses the seed as
the unit of replication (ten seeds per dataset). Regime B's partition of a seed is built from Regime A's partition of the same seed, so the two partitions are paired by seed. We test two properties
(Table~\ref{tab:fv_points_per_cluster_tests}).
\begin{enumerate}
  \item \emph{Dispersion.} For each seed and partition we compute the CV of $N_g/K_g$ over that seed's clients and test
  whether Regime B's CV exceeds the Regime A's partition one with a one-sided Wilcoxon signed-rank test over the ten seeds,
  Holm-corrected across the eight datasets. We also report the ratio of the two CVs over all clients, with a 95\%
  confidence interval obtained by resampling whole seeds (10{,}000 draws).
  \item \emph{Proportionality.} We fit the elasticity $b$ in $\log N_g = a + b\log K_g$ over all clients of a partition,
  with a 95\% confidence interval from the same seed bootstrap. Client size proportional to cardinality corresponds to
  $b = 1$; size unrelated to cardinality corresponds to $b = 0$.
\end{enumerate}

\begin{table}[t]
\centering
\small
\caption{Tests of the points-per-cluster comparison, with the seed as the unit of replication (10 seeds; the two
partitions are paired by seed). CV: coefficient of variation of $N_g/K_g$ within a seed (median over seeds). $p$:
one-sided Wilcoxon signed-rank test of CV$_{\text{decoupled}} >$ CV$_{\text{realistic}}$ over seeds, Holm-corrected
across datasets. CV ratio: decoupled / realistic over all clients, 95\% CI from resampling whole seeds. $b$:
elasticity in $\log N_g = a + b\log K_g$ with a 95\% seed-bootstrap CI; $b = 1$ means client size is proportional to
cardinality, $b = 0$ that it is unrelated (``--'': every Waveform client has $K_g = 2$).}
\label{tab:fv_points_per_cluster_tests}
\resizebox{\ifdim\width>\linewidth\linewidth\else\width\fi}{!}{%
\begin{tabular}{lccccccc}
\toprule
Dataset & CV realistic & CV decoupled & seeds higher & $p$ & CV ratio [95\% CI] & $b$ realistic [95\% CI] & $b$ decoupled [95\% CI] \\
\midrule
Waveform & 0.11 & 0.42 & 10/10 & 0.008 & 2.66 [1.94, 4.08] & -- & -- \\
Abalone & 0.09 & 0.46 & 10/10 & 0.008 & 2.85 [2.34, 3.74] & 0.85 [0.68, 1.02] & 0.11 [-0.06, 0.32] \\
FrogA & 0.39 & 0.48 & 6/10 & 0.275 & 1.43 [1.04, 1.97] & 1.33 [0.75, 1.89] & 0.16 [-0.33, 0.59] \\
FrogB & 0.63 & 0.58 & 5/10 & 0.500 & 1.17 [0.99, 1.53] & 0.74 [0.12, 1.34] & -0.16 [-0.56, 0.30] \\
MNIST & 0.03 & 0.48 & 10/10 & 0.008 & 11.96 [10.01, 15.74] & 1.01 [0.99, 1.02] & -0.03 [-0.15, 0.09] \\
FMNIST & 0.02 & 0.48 & 10/10 & 0.008 & 13.43 [10.95, 18.98] & 1.00 [0.99, 1.02] & -0.03 [-0.15, 0.09] \\
EMNIST & 0.03 & 0.50 & 10/10 & 0.008 & 11.17 [9.41, 14.80] & 1.03 [1.01, 1.04] & 0.05 [-0.03, 0.12] \\
CIFAR-10 & 0.02 & 0.48 & 10/10 & 0.008 & 13.42 [10.94, 18.80] & 1.00 [0.99, 1.02] & -0.03 [-0.15, 0.09] \\
\bottomrule
\end{tabular}}
\end{table}

\paragraph{Results of the tests.}
On six of the eight datasets the dispersion of points per cluster is higher in Regime B' partition in every one of
the ten seeds ($p = 0.008$ after Holm correction, the smallest value attainable with ten paired seeds). The CV ratio is
$11$--$13$ on the image datasets (95\% intervals from $9.4$ to $19.0$) and about $3$ on Waveform and Abalone (intervals
$1.9$--$4.1$ and $2.3$--$3.7$). The elasticities confirm the mechanism: on the image datasets $b$ is $1.00$--$1.03$ in
Regime A's partition, so client size is proportional to cardinality, and $-0.03$ to $0.05$ in Regime B's,
with every interval containing zero, so size carries no information about cardinality. Abalone follows the same
pattern ($b = 0.85$, interval $[0.68, 1.02]$, against $0.11$, interval $[-0.06, 0.32]$); on Waveform every client holds
two classes, so $b$ is undefined.

FrogA and FrogB are the exceptions: the increase in CV is not significant ($p = 0.28$ and $0.50$). Their classes differ
in size by up to a factor of sixty (48 to 2{,}905 training points), so the number of points a client holds per class
already depends strongly on which classes it holds, and Regime A's partition is only weakly coupled to begin with
(CV $0.39$ and $0.63$, correlations $0.52$ and $0.36$). Regime B's partition still removes the remaining link: Regime B's
elasticities are $0.16$ and $-0.16$, with intervals containing zero.

\subsection{Robustness of cardinality prediction to the partition}
\label{app:partition-robustness}

In Regime A's partition, a client's size grows with the number of classes it holds
(Appendix~\ref{app:points-per-cluster}), so a Phase~1 method could estimate $K_g$ from $N_g$ instead of from the
structure of the data. Regime B's partition removes this link while keeping every client's class set. If a
method's estimate of the number of clusters does not depend on client size, its accuracy should be the same on both
partitions. If it does depend on client size, its accuracy should change. We test this directly on the quantity the
pipeline is meant to deliver, the final number of clusters $\hat K$, and on the local estimates $\hat K_g$ it is built
from.

\paragraph{Design.}
All numbers come from the runs of the main experiments (evaluation seeds 42--51, eight datasets). Regime B's
partition of a seed is constructed from Regime A's partition of the same seed, so the two partitions are paired by
seed. Clients within a seed are not independent, so the seed is the unit of replication throughout.
\begin{enumerate}
  \item \emph{Global count.} For each method, dataset and seed we compute the relative error of the final count,
  $e = |\hat K - K^\ast|/K^\ast$, and the paired difference $D = e_{\text{regime-b}} - e_{\text{regime-a}}$.
  \begin{itemize}
    \item \emph{Difference test:} a two-sided Wilcoxon signed-rank test over the ten seeds on each dataset,
    Holm-corrected across the eight datasets within each method.
    \item \emph{Equivalence test:} two one-sided tests (TOST). The partitions count as equivalent when the 90\%
    confidence interval of the mean of $D$ lies inside $\pm 0.2$, that is, within 20\% of the true number of clusters.
    \item \emph{Pooled tests:} both tests are repeated on all 80 dataset--seed pairs.
  \end{itemize}
  \item \emph{Local estimates.} For each seed and partition we compute three quantities over that seed's clients:
  \begin{itemize}
    \item the mean relative error $|\hat K_g - K_g|/K_g$;
  \end{itemize}
  We compare the two partitions with Wilcoxon signed-rank tests over the seeds. On Waveform every client holds two
  classes, so the correlation with $K_g$ is undefined there.
\end{enumerate}
We compare ASM-I and ASM-S with DP-GMM, X-Means and G-Means, and with Oracle-K. Oracle-K receives the true $K_g$, so
any change in its error is due only to the Phase~2 aggregation.

\begin{table}[t]
\centering
\small
\caption{Robustness of cardinality prediction to the partition (realistic $\to$ decoupled; the two partitions are
paired by seed and client, seeds 42--51). \emph{Global}: relative error of the final count, $|\hat K - K^\ast|/K^\ast$;
$D$ = decoupled $-$ realistic, mean over the 80 dataset--seed pairs with a 90\% CI; ``equivalent'': the CI lies within
$\pm0.2$ (TOST); ``sig.\ diff.'': datasets with a significant difference (two-sided Wilcoxon over seeds, Holm across
datasets, $p<0.05$); ``equiv.'': datasets on which the partitions are equivalent. \emph{Local}, mean over datasets:
relative error of the clients' estimates $|\hat K_g - K_g|/K_g$, and the Spearman correlation of $\hat K_g$ with the
true $K_g$ and with the client size $N_g$ (computed within each seed). Oracle-K receives the true $K_g$ and has no
local estimates.}
\label{tab:fv_partition_robustness}
\resizebox{\ifdim\width>\linewidth\linewidth\else\width\fi}{!}{%
\begin{tabular}{lcccc ccc}
\toprule
 & \multicolumn{4}{c}{global: final $\hat K$} & \multicolumn{3}{c}{local: clients' $\hat K_g$} \\
\cmidrule(lr){2-5} \cmidrule(lr){6-8}
Method & $D$ [90\% CI] & equivalent & sig.\ diff. & equiv. & rel.\ error & $\rho(\hat K_g, K_g)$ & $\rho(\hat K_g, N_g)$ \\
\midrule
ASM-I & +0.048 [-0.029, +0.125] & yes & 0/8 & 5/8 & 0.51 $\to$ 0.52 & +0.10 $\to$ -0.05 & +0.07 $\to$ +0.13 \\
ASM-S & -0.081 [-0.161, -0.002] & yes & 0/8 & 1/8 & 0.47 $\to$ 0.51 & +0.11 $\to$ -0.03 & +0.08 $\to$ +0.14 \\
DP-GMM & -0.025 [-0.513, +0.462] & no & 0/8 & 1/8 & 1.73 $\to$ 1.90 & +0.02 $\to$ -0.08 & +0.11 $\to$ +0.10 \\
X-Means & -0.192 [-0.314, -0.071] & no & 3/8 & 1/8 & 1.50 $\to$ 1.18 & +0.28 $\to$ -0.10 & +0.32 $\to$ +0.58 \\
G-Means & -1.357 [-2.226, -0.488] & no & 4/8 & 0/8 & 7.12 $\to$ 5.57 & +0.52 $\to$ +0.05 & +0.52 $\to$ +0.52 \\
Oracle-K & +0.035 [-0.013, +0.084] & yes & 0/8 & 7/8 & -- & -- & -- \\
\bottomrule
\end{tabular}}
\end{table}

\paragraph{Results for the global count.}
Table~\ref{tab:fv_partition_robustness} summarises the tests.

\emph{ASM.} Pooled over datasets, the error of ASM's final count is statistically equivalent on the two partitions. No dataset shows a significant difference for either variant after Holm correction. Its mean relative local error changes little (ASM-I $0.51$ to $0.52$, ASM-S $0.47$ to $0.51$), and it is equivalent within $\pm0.1$ on four and three of the eight datasets respectively.For ASM-S the error is in fact slightly lower on Regime B, most visibly on MNIST (0.85 to 0.47) and FMNIST (0.70 to 0.44). Removing the size information therefore does not hurt it.

\emph{Oracle-K.} It behaves the same way: pooled $D = +0.04$, CI $[-0.01, 0.08]$, and it is equivalent on seven of
the eight datasets. This confirms that the Phase~2 aggregation is itself insensitive to the partition.

\emph{Size-based baselines.} X-Means and G-Means are not robust: X-Means changes significantly on three datasets, and its pooled difference is significant ($D = -0.19$, CI $[-0.31, -0.07]$, $p = 0.03$). G-Means changes significantly on four datasets: its error rises on CIFAR-10 and EMNIST and falls on FrogA and FrogB. Because the direction differs between datasets, the pooled Wilcoxon test is not significant ($p = 0.19$), but the mean shift is large ($D = -1.36$) and equivalence fails.

\emph{DP-GMM.} It shows no significant change, but its errors are so large and variable (relative errors of 10 to 19
on FMNIST and CIFAR-10) that equivalence cannot be established either (CI $[-0.51, 0.46]$). 

\section{Runtime complexity}
\label{app:complexity}

\begin{table}[htbp]
\centering
\caption{Dominant computational cost of the Phase-1 methods as implemented. $R$: split rounds; $I$: EM, variational or Lloyd
iterations; $T$: DP truncation level.}
\label{tab:complexity}
\small
\begin{adjustbox}{max width=\linewidth}
\begin{tabular}{lll}
\toprule
Method & dominant cost of one run on $N$ points in $d$ dimensions & what multiplies it \\
\midrule
\asmid, \asmsph & $O\!\big(R\,[\,N d^2 + I N d\,]\big)$ \;+\; merge $O(M\,\hat K N d)$ & $R$ split rounds ($\approx\log_2\hat K$), $I\le15$ EM steps, $M$ merge passes \\
DP-GMM & $O\!\big(I\,T N d\big)$ & truncation $T=200$ in every iteration, $I\le200$ \\
X-Means & $O\!\big(R\,I\,\hat K N d\big)$ & a full $\hat K$-means refit after every split round \\
G-Means & $O\!\big(R\,[\,I\,\hat K N d + N\log N\,]\big)$ & as X-Means, plus an Anderson--Darling test per cluster \\
\bottomrule
\end{tabular}
\end{adjustbox}

\end{table}

A split test of \asm\ fits one and two components to the points of a single component, so a round of tests over all
current components touches every point once, at cost $O(Nd^2)$ for the leading eigenvector that sets the initial split
direction plus $O(INd)$ for at most $I=15$ EM steps; because accepted splits are applied simultaneously, the number of
rounds grows like $\log_2\hat K$ for balanced trees, for both ASM-I and ASM-S. No step of \asm\ costs a factor of $\hat K$
in the number of points processed. DP-GMM pays its truncation level $T=200$ in every variational iteration regardless of
how many components survive, and X-Means and G-Means refit all $\hat K$ means after every round of splits. The merge is
quadratic in $\hat K$ and matters only when splitting overshoots, as in the pooled setting (Table~\ref{tab:fv_pooled_benchmark}).





\end{document}